\documentclass[manuscript,screen]{acmart}
\usepackage{colortbl}

\usepackage{graphicx}
\usepackage{subfigure}
\usepackage{pdfpages}
\usepackage{multirow} 
\usepackage{textcomp}
\usepackage{booktabs} 
\usepackage{float}
\usepackage{pifont} 

\AtBeginDocument{%
  }

\acmISBN{978-1-4503-XXXX-X/2018/06}

\begin{document}
\newcommand{\xmark}{\ding{53}}
\title[PE-CLIP: A Parameter-Efficient Fine-Tuning of VLMS for DFER]{PE-CLIP: A Parameter-Efficient Fine-Tuning of Vision Language Models for Dynamic Facial Expression Recognition}

\author{Ibtissam Saadi}
\orcid{0009-0002-4231-4246}
\authornotemark[1]
\email{ibtissam.ssadi@b-tu.de}
\affiliation{%
  \institution{Faculty of Graphical Systems, Univ. BTU Cottbus-Senftenberg}
  \city{Cottbus}
  \country{Germany}
}

\author{Abdenour Hadid}
\orcid{0000-0001-9092-735X}
\affiliation{%
 \institution{Sorbonne Center for Artificial Intelligence, Sorbonne University Abu Dhabi}
  \city{Abu Dhabi}
  \country{UAE}}
\email{abdenour.hadid@ieee.org}

\author{Douglas W. Cunningham}
\orcid{0000-0003-1419-2552}
\affiliation{%
  \institution{Faculty of Graphical Systems, Univ. BTU Cottbus-Senftenberg}
  \city{Cottbus}
  \country{Germany}
}
\email{douglas.cunningham@b-tu.de}

\author{Abdelmalik Taleb-Ahmed}
\orcid{0000-0001-7218-3799}
\email{abdelmalik.taleb-ahmed@uphf.fr} 
\author{Yassin El Hillali}
\email{yassin.elHillali@uphf.fr}
\orcid{0000-0002-3980-9902}
\affiliation{%
 \institution{Laboratory of IEMN, CNRS, Centrale Lille, UMR 8520, Univ. Polytechnique Hauts-de-France, F-59313}
 \city{Valenciennes}
 \country{France}}





\renewcommand{\shortauthors}{Saadi {\it et al.}}

\begin{abstract}
  The emergence of Vision-Language Models (VLMs) like CLIP (Contrastive Language-Image Pretraining) provides appealing solutions to various vision problems including Dynamic Facial Expression Recognition (DFER). However, most of the proposed approaches face major challenges, particularly related to inefficient full fine-tuning of the encoders and the complexity of the models. Moreover, some of the proposed methods seem to struggle with suboptimal performance due to (i) poor alignment between textual and visual representations, and (ii) ineffective temporal modeling. To address these challenges, we propose PE-CLIP, a parameter-efficient fine-tuning (PEFT) framework that elegantly adapts CLIP for dynamic facial expression recognition, requiring significantly reduced number of trainable parameters while maintaining high accuracy. At its core, to enhance efficiency and performance, PE-CLIP introduces two specialized adapters namely a Temporal Dynamic Adapter (TDA) and a Shared Adapter (ShA). The temporal dynamic adapter is a GRU-based module with a dynamic scaling mechanism, capturing sequential dependencies while adaptively modulating the contribution of each temporal feature to emphasize the most informative ones while mitigating irrelevant variations. The shared adapter is a lightweight adapter 
  refine representations within both textual and visual encoders, ensuring consistent feature processing while maintaining parameter efficiency. Additionally, we leverage Multi-modal Prompt Learning (MaPLe), which introduces learnable prompts to both visual and action unit-based textual description inputs, further improving the semantic alignment between modalities and enabling the efficient adaptation of CLIP for dynamic tasks. We evaluate our proposed PE-CLIP on two benchmark datasets, namely DFEW and FERV39K, achieving competitive performance compared to state-of-the-art methods while requiring fewer trainable parameters. By striking an optimal balance between parameter efficiency and performance, PE-CLIP sets a new benchmark in resource-efficient DFER. The source code of the proposed PE-CLIP will be publicly available at \href{https://github.com/Ibtissam-SAADI/PE-CLIP}{https://github.com/Ibtissam-SAADI/PE-CLIP}.
\end{abstract}



\maketitle

\section{Introduction}
\label{sec:intro}


Early research on Facial Expression Recognition (FER) mainly focused on static FER (SFER), which analyzes individual images to identify facial expressions~\cite{chattopadhyay2020facial,saadi2023driver}. 
Fundamentally, static images may fail capturing the temporal dynamics of facial movements - an essential factor in understanding human emotions. Dynamic FER (DFER), on the other hand, utilizes video sequences to analyze temporal changes in facial expressions, capturing the progression of facial dynamics over time. This enables a more nuanced understanding of emotional states, making DFER better suited for real-world applications. In particular, the availability of in-the-wild DFER datasets, such as DFEW~\cite{jiang2020dfew}, FERV39k~\cite{wang2022ferv39k}, and MAFW~\cite{liu2022mafw}, which closely reflect real-world conditions, has advanced the field but also presents inherent challenges. These include occlusions, motion blur, non-frontal faces, and low-resolution images, all of which hinder the accurate analysis of dynamic facial features. 
Researchers have explored various approaches to address these challenges. Early methods used Convolutional Neural Networks (CNNs) for spatial feature extraction, combined with Recurrent Neural Networks (RNNs) for temporal modeling (CNN-RNN)~\cite{sun2020multi,9102419}. Other approaches leveraged 3D Convolutional Neural Networks (3D CNNs)~\cite{10204167,10.1145/2993148.2997632,khanna2024enhanced} to jointly model spatial and temporal information, providing a unified representation of facial dynamics. Transformer-based architectures~\cite{zhao2021former,li2023intensity, xia2023hit} have also been proposed to leverage global attention and long-range dependency modeling to improve spatio-temporal feature learning. Despite of the notable advances in the proposed works, some challenges still remain. The primary issue lies in the effective modeling of the complex temporal correlations within facial expression sequences, especially in the presence of subtle and hidden expressions. Furthermore, the scarcity of the training data in existing DFER datasets limits the development of deep models. To tackle these challenges, recent works like MAE-DFER~\cite{sun2023mae} and MARLIN~\cite{cai2023marlin} have utilized large-scale self-supervised pretraining on unlabeled facial videos to enhance learning from limited data and improve temporal modeling through techniques such as mask modeling. However, despite achieving promising results, such methods usually involve high computational costs due to the heavy pretraining and inadequate representation of subtle temporal variations in facial expressions.

More recently, researchers have begun exploring vision-language models due to their exceptional generalization capabilities across various downstream tasks. These models, such as Contrastive Language-Image Pretraining 
(CLIP) ~\cite{radford2021learning}, represent a significant shift from vision-only approaches by integrating textual and visual modalities. By leveraging the alignment between visual and textual input, vision-language models have shown great promise in enhancing the understanding of facial expressions. Recent works like CLIPER~\cite{li2024cliper} and DFER-CLIP~\cite{zhao2023prompting} have adapted CLIP for DFER, introducing textual prompts that provide additional semantic information to visual features and achieving notable improvements in performance. However, while these CLIP-based methods have made significant progress, they also face new challenges. Many of the approaches require full fine-tuning of the image encoder, leading to high computational costs and resource-intensive training. This limits their practicality, especially in resource-constrained scenarios. Furthermore, temporal modeling in these methods often relies on temporal average pooling or single-layer Transformers, which struggle to capture complex short- and long-term dependencies crucial for dynamic facial expression analysis. 
Additionally, while textual prompts provide valuable semantic context, existing methods often fail to fully exploit enriched and task-specific textual descriptions tailored to dynamic facial sequences. This limitation reduces the alignment between visual and textual modalities, leading to suboptimal performance in capturing dynamic facial expressions. To mitigate the high computational cost associated with fine-tuning large pre-trained models, recent efforts have explored parameter-efficient fine-tuning (PEFT) strategies~\cite{houlsby2019parameter, chen2022adaptformer}. 
However, these approaches primarily focus on single-modality adaptation and employ simplistic temporal modeling techniques. They lack dedicated temporal adapters specifically designed for CLIP in DFER. 

From the observations above, we believe that there is a need for methods that leverage advancements in PEFT while incorporating specialized temporal modeling mechanisms. Based on this motivation, we propose PE-CLIP, a novel parameter-efficient adaptation framework designed to adapt CLIP for DFER. PE-CLIP introduces specialized components to efficiently model temporal dynamics, improve alignment between the two modalities' representations, and achieve competitive performance with reduced computational cost. Unlike prior works that rely on computationally expensive architectures or inadequate temporal modeling, PE-CLIP integrates adapters and prompt learning to provide a more efficient and robust solution.


\begin{figure}[!b]
      \centering
      \includegraphics[width=0.8\linewidth]{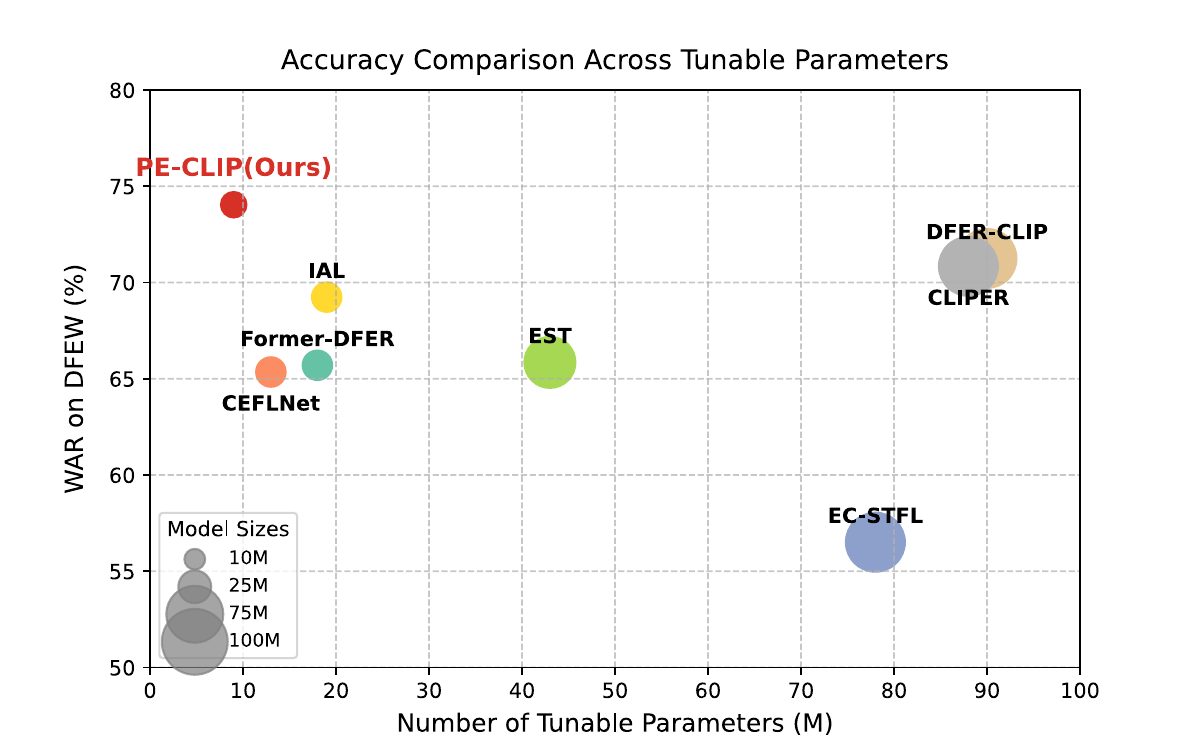}
    \caption{Performance comparison of dynamic facial expression recognition on the DFEW dataset. This chart highlights the trade-off between the number of tunable parameters (in millions) and model accuracy (measured as weighted average recall, WAR). The bubble size represents the model size, indicating the total number of trainable parameters. Our proposed framework, PE-CLIP (highlighted in red), achieves superior performance with significantly fewer tunable parameters ($<$6\% of the whole model's parameters) compared to state-of-the-art methods. Models included in the comparison are IAL~\cite{lee2023frame}, CEFLNet~\cite{liu2022clip} , EC-STFL~\cite{jiang2020dfew}, Former-DFER~\cite{zhao2021former}, EST~\cite{liu2023expression}, DFER-CLIP~\cite{zhao2023prompting}, and CLIPER~\cite{li2024cliper}.}
       \label{fig:pip}
\end{figure}

Our proposed framework comprises two  adapter modules, namely the Temporal Dynamic Adapter (TDA) and a Shared Adapter (ShA). The temporal dynamic adapter utilizes a GRU-based architecture with a dynamic scaling mechanism to capture both short- and long-term temporal dependencies while adaptively emphasizing the most informative temporal features in video sequences. The shared adapter is a lightweight vanilla adapter that efficiently refines features across both textual and visual modalities. Additionally, PE-CLIP leverages Multi-modal Prompt Learning (MaPLe)~\cite{ekman1997face} to adapt both textual and visual inputs by introducing a learnable prompt to both inputs. The textual supervision is enriched using facial action units (AUs)~\cite{ekman1997face} descriptions, which provide semantically meaningful and expression-specific representations that enhance alignment with dynamic visual features. PE-CLIP effectively balances computational efficiency and model performance, as illustrated in Figure~\ref{fig:pip}. PE-CLIP indeed achieves superior accuracy while reducing tunable parameters, requiring less than 6\% of the total model parameters. Comprehensive evaluation on benchmark datasets, including DFEW and FERV39K, validates the robustness and effectiveness of PE-CLIP for dynamic facial expression recognition.

The main contributions of this work can be summarized as follows: 
\begin{itemize} 
\item We propose PE-CLIP, a new parameter-efficient framework designed for dynamic facial expression recognition. By leveraging lightweight adapter modules and multi-modal prompt learning. PE-CLIP enhances visual-textual alignment in CLIP while maintaining computational efficiency.

\item We introduce a Temporal Dynamic Adapter (TDA) which is a GRU-based module
with a dynamic scaling mechanism, enabling adaptive emphasis on significant temporal variations. 

\item We introduce a Shared
Adapter (ShA) which is a vanilla adapter applied across both the
textual and visual modalities to improve representations, providing consistent feature refinement while maintaining parameter efficiency within frozen CLIP encoders.

\item We leverage Multi-modal prompt Learning (MaPLe) to adapt both textual and visual inputs. For textual input, we enrich prompts with Action Unit (AU) descriptions, providing semantically rich and contextually relevant prompts for expressions. This enhancement strengthens semantic alignment between textual and visual representations, ensuring more expressive feature adaptation for DFER. 

\item We conduct extensive experiments and comparison with state-of-the-art on benchmark datasets, validating the effectiveness of our proposed framework.

\end{itemize}

The rest of this article is structured as follows: Section~\ref{sec:rwork} provides an overview of related works, highlighting advancements in DFER, 
discussing CLIP model applications, and parameter-efficient transfer learning techniques. Section~\ref{sec:met} introduces the proposed PE-CLIP framework, detailing its architecture and core components. Section~\ref{sec:exp} presents the extensive experimental analysis, including the experimental data, the implementation details, the obtained results, the comparison against state-of-the-art and a thorough ablation study to gain insight into the contribution of each part of the proposed framework. Finally, Section~\ref{sec:clc} draws conclusions and highlights potential future research directions.


\section{Related Work}
\label{sec:rwork}

Dynamic facial expression recognition aims to understand and classify facial expressions in videos by capturing both spatial and temporal dynamics. Early approaches primarily relied on handcrafted features, such as Local Binary Patterns (LBP)~\cite{LBP}, evaluated on lab-controlled datasets~\cite{zhang2014bp4d},~\cite{livingstone2018ryerson}. While these methods were effective in controlled settings, their limited ability to generalize to real-world scenarios restricted their applicability. With the advent of deep learning and the availability of large-scale in-the-wild datasets~\cite{liu2022mafw},~\cite{jiang2020dfew},~\cite{wang2022ferv39k}, data-driven techniques have become the foundation for advancing DFER research. Recent deep learning-based DFER methods can be broadly categorized into three main categories. The first category is CNN-RNN-based methods which typically utilizes CNNs to extract spatial facial features from each frame, followed by RNNs to model temporal dependencies across sequences. Commonly used CNN backbones include VGG~\cite{simonyan2014very} and ResNet~\cite{he2016deep}, while LSTM~\cite{hochreiter1997long} and GRU~\cite{chung2014empirical} were widely used for the temporal modeling. For instance, the authors in~\cite{sun2020multi} enhanced the LSTM's ability for long-term contextual modeling by augmenting it with a self-attention mechanism. The work in ~\cite{9102419} introduced a multi-modal recurrent attention network designed to capture temporal features by integrating multi-dimensional information. More recently, a hybrid CNN-RNN architecture using time distributed layers for sequential video processing was proposed~\cite{manalu2024detection}, demonstrating improved efficiency. The second category of methods, 3D CNN-based methods, involves leveraging 3D CNNs to jointly capture spatial and temporal information~\cite{7410867}. Early works in this category such as~\cite{10.1145/2993148.2997632} employed 3D-CNNs to model video appearances and motions. Similarly, the Multi-3D Dynamic Facial Expression Learning (M3DFEL) model introduced in~\cite{10204167} generated 3D instances to model short-term temporal relationships and utilized 3D-CNNs for feature extraction. In another work~\cite{khanna2024enhanced}, the authors utilized two 3D-CNN architectures: a 3D ResNet model to extract feature vectors from video sequences and a 3D DenseNet model for emotion classification. The third and emerging category of methods are based on transformers. These methods have gained attention due to the great success of transformer models~\cite{dosovitskiy2020image}. Transformers have been utilized as spatial, temporal, or spatiotemporal feature extractors for DFER. For example, the work in \cite{zhao2021former} introduced Former-DFER, which combined convolutional spatial transformers with temporal transformers to extract robust spatial and temporal facial features. In another work \cite{zhang2023transformer}, a multimodal framework based on Transformers was proposed, integrating audio, video, and text modalities for more comprehensive emotional feature extraction. In \cite{li2023intensity}, the authors developed a global convolution-attention block that rescaled feature map channels and used temporal transformers to capture temporal relationships. More recently, Jin {\it et al.} proposed a transformer-embedded spectral-based graph convolution network that interacts with complex relationships among facial regions of interest (ROIs) for FER \cite{jin2024transformer}. 

Despite the notable progress, most of the methods reported above face some limitations. For instance, CNN-RNN-based approaches seem to struggle with effectively capturing the temporal correlations, particularly when dealing with subtle or long-term dependencies in dynamic facial expressions. While 3D CNNs jointly model spatial and temporal information, they suffer from the high computational costs and the scalability issues when applied to deeper architectures. Transformer-based methods, though effective in leveraging global attention, are computationally intensive and requiring large training datasets. 
To cope with some of these challenges, researchers started looking at vision-language models, particularly  CLIP~\cite{radford2021learning}, due to their adaptability across a range of tasks~\cite{lin2024rethinking,ghosh2024clipsyntel,10.1007/978-3-031-80856-2_6,yu2024tf}. CLIP offers the unique capability to integrate visual and textual modalities, which has been applied to DFER with promising results. Several recent studies have proposed adaptations of CLIP for DFER. For instance, the authors in \cite{zhao2023prompting} introduced DFER-CLIP, and extended CLIP by incorporating a transformer-based temporal module, which improved the temporal modeling capabilities but required full fine-tuning of both the CLIP image encoder and the additional module, leading to inefficiencies in training and computational cost. similarly, CLIPER~\cite{li2024cliper} was introduced as a two-stage training paradigm to adapt CLIP for DFER but relied heavily on temporal average pooling to aggregate video features, limiting its ability to capture nuanced temporal dynamics. 
Recently, the authors in \cite{10744485} proposed a novel visual language model that captures and aggregates dynamic features of expressions in semantic supervision via Inter-Frame Interaction Transformer (Inter-FIT) and Multi-Scale Temporal Aggregation (MSTA). While these methods demonstrate the potential of CLIP’s pre-trained capabilities, they remain computationally demanding and often lack effective temporal modeling and robust semantic alignment.

As the scale of pre-trained models continues to grow, the computational demands for fully fine-tuning these models have become increasingly prohibitive. To mitigate this issue, Parameter-Efficient Transfer Learning (PETL) techniques~\cite{houlsby2019parameter} have been developed, initially for natural language processing (NLP) tasks, and later adapted for computer vision. PETL methods aim to significantly reduce the number of trainable parameters, thereby lowering computational costs while maintaining or enhancing model performance. These techniques are broadly classified into two categories: adapter learning, which involves inserting lightweight modules into model layers for incremental fine-tuning~\cite{chen2022adaptformer, yang2023aim, xin2024vmt}, and prompt learning~\cite{jia2022visual, xing2023dual, yang2024dgl}, which incorporates learnable tokens to adapt the input for specific tasks. In the context of DFER, few studies have explored PETL techniques, particularly for adapting static facial expression models to dynamic video-based recognition. For instance, FE-Adapter~\cite{gowda2024fe} introduced a single adapter leveraging dynamic dilated convolutions to improve video emotion recognition. Similarly, S2D~\cite{chen2024static} proposed a learnable Temporal Modeling Adapter (TMA) based on multi-head self-attention, inserted between adjacent transformer blocks within a residual structure. However, these approaches primarily focus on adapting the vision encoder rather than multi-modal alignment. 
In contrast, our method explores both adapters and prompt learning to enhance efficiency and performance. Specifically, we introduce the Temporal Dynamic Adapter (TDA), a GRU-based module designed for effective temporal modeling, and the Shared Adapter (ShA) to refine features while ensuring consistent processing across modalities. Additionally, we leverage MaPLe, incorporating enriched textual descriptions to further improve vision-language alignment, ultimately striking an optimal balance between parameter efficiency and recognition performance in DFER.


\section{Proposed Framework}
\label{sec:met}

\begin{figure*}[!b]
      \centering
      \includegraphics[width=\linewidth]{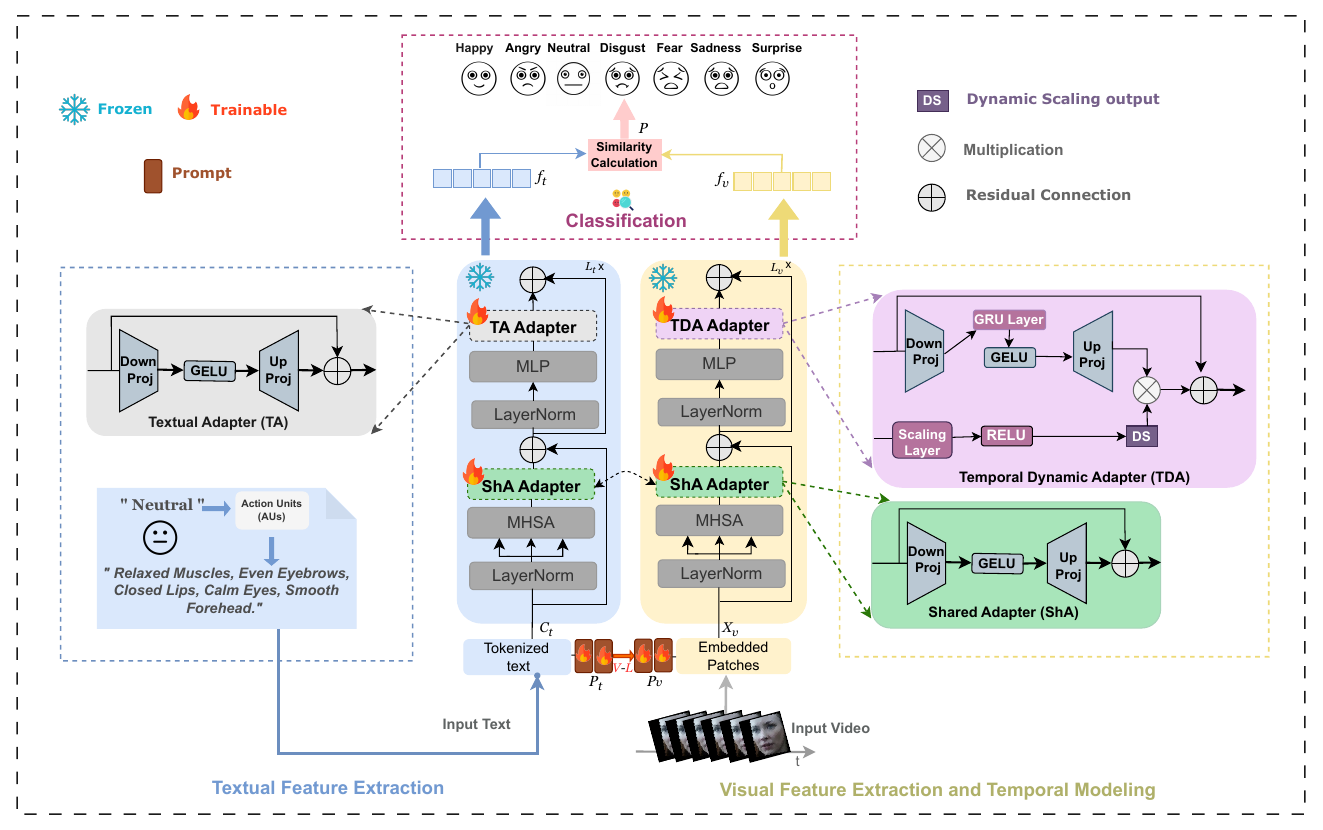}
      \caption{Overall architecture of the proposed PE-CLIP model. The model takes tokenized AU-based textual descriptions $\mathbf{\textit{C}}_{t}$ and embedded image sequences $\mathbf{\textit{X}}_{t}$, enriched with MaPLe learnable tokens as inputs. These representations are processed through CLIP encoders, where the Shared Adapter (ShA) improves the representation learning across textual-visual modalities. Additionally, the Temporal Dynamic Adapter (TDA) with dynamic scaling, captures the key temporal dependencies in the vision branch, while the Textual Adapter (TA) enhances the textual representations. The resulting visual ($\mathbf{\textit{f}}_{v}$) and textual ($\mathbf{\textit{f}}_{t}$) embeddings are mapped into CLIP’s shared space, where classification is performed via cosine similarity to associate expressions with their corresponding labels.}
      \label{fig:arch}
  \end{figure*}

This section details the proposed PE-CLIP framework, which builds on the pre-trained CLIP model~\cite{radford2021learning} and adapts it for DFER using parameter-efficient fine-tuning. The core components of our methodology are described in detail. Section~\ref{sec:pre} introduces the foundational concepts of the CLIP model and provides an overview of the PE-CLIP framework. Sec~\ref{sec:vs} focuses on visual feature extraction and temporal modeling. Section~\ref{sec:ts} explains textual prompt tuning, and Section~\ref{sec:class} presents classification via cosine similarity. The complete pipeline of the framework is illustrated in Figure~\ref{fig:arch}.
  
\subsection{Overview of PE-CLIP}
\label{sec:pre}
CLIP~\cite{radford2021learning} is a vision-language model designed to learn a shared embedding space for images and text. It comprises two separate encoders: an image encoder and a text encoder, independently extracting high-dimensional embeddings from visual and textual inputs. During training, CLIP optimizes a contrastive loss function to align image and text embeddings by maximizing the similarity of matching pairs while minimizing that of mismatched pairs. For classification tasks, CLIP projects visual features $v$
and textual features $t_k$(representing class labels $k$) into a shared embedding space. The prediction of an image’s class using the pre-trained CLIP model can be formulated as:
\begin{equation}
    P(k|v) = \frac{\exp(\text{cos}(v, t_k)/\tau)}{\sum_{j=1}^{C} \exp(\text{cos}(v, t_j)/\tau},
    \label{eq:cosine_similarity}
\end{equation}\\
where $\text{cos}(v, t_k)$ denotes the cosine similarity between the visual and textual embeddings, $C$ is the total number of classes, and $\tau$ is a learnable temperature parameter. This shared embedding space allows CLIP to perform zero-shot classification by matching image embeddings to textual embeddings derived from class names or prompts. In this work, we effectively and efficiently adapt CLIP for DFER using adapters and prompt tuning.
As illustrated in Figure~\ref{fig:pip}, the proposed PE-CLIP framework comprises three main components: visual feature extraction and temporal modeling, textual feature extraction, and classification. For the visual feature extraction part, a facial expression video sequence consisting of \(T\) frames is processed using the frozen CLIP visual encoder (ViT-B/16). The visual embeddings for each frame are refined by lightweight adapters, specifically the Shared Adapter (ShA), applied after the Multi-Head Self-Attention (MHSA) layers to efficiently refine the extracted features, and the Temporal Dynamic Adapter (TDA), applied after the Multi-Layer Perceptron (MLP) layers to model temporal dependencies across frames. In the textual feature extraction part, fine-grained prompts derived from facial action units (AUs) based on the Facial Action Coding System (FACS)~\cite{ekman1997face} are used to replace class names, providing richer semantic information as established in prior work~\cite{ren2024facial}. These prompts are further refined by introducing learnable tokens via the Multi-modal Prompt Learning (MaPLe)~\cite{khattak2023maple} method, enabling the model to better capture nuanced relationships between textual descriptions and dynamic facial expressions. The textual embeddings are then processed using the ShA adapter to refine extracted features consistently and the Textual Adapter (TA) to enhance the representation quality before alignment with the visual features. Finally, in the classification part, the refined visual and textual embeddings are projected into the shared embedding space, where cosine similarity determines the class label. Further details on these components are provided in the subsequent sections.
\subsection{Visual Feature Extraction and Temporal Modeling}
\label{sec:vs}
A key challenge in adapting the pre-trained CLIP model for DFER is achieving effective temporal modeling across video frames while ensuring parameter efficiency. Unlike static image recognition tasks or general video recognition, DFER requires modeling subtle short- and long-term dependencies in facial expressions. Although CLIP’s pre-trained Vision Transformer (ViT) encoder~\cite{alexey2020image} excels in extracting spatial representations, it lacks mechanisms to directly model the temporal dynamics critical for DFER. To address this limitation, we extend the frozen CLIP visual encoder with lightweight, parameter-efficient adapters specifically designed to refine spatial features and capture temporal dependencies. The process begins with frame-level feature extraction from a facial expression video sequence $\mathbf{X} \in \mathbb{R}^{T \times H \times W \times 3}$, where $T$ is the temporal length, and $H \times W$ denotes the spatial resolution of each frame. Each frame $\mathbf{X}_t \in \mathbb{R}^{H \times W \times 3}$ is divided into non-overlapping patches of size $P \times P$, resulting in $M = (H \cdot W) / P^2$ patches per frame. These patches are flattened and projected into a $d$-dimensional latent space via a patch embedding layer, producing patch embeddings ${\mathbf{X}{v,i}}_{i=1}^M \in \mathbb{R}^d$. The learnable prompts $\mathbf{P}_v$ are prepended to these patch embeddings and then are processed within the frozen CLIP visual encoder through a series of transformer layers comprising the MHSA layer and MLPs. After processing, the class token is used to derive a frame-level feature embedding $\mathbf{v}_t \in \mathbb{R}^d$. Consequently, a sequence of frame-level embeddings $\mathbf{V} = {\mathbf{v}_1, \mathbf{v}_2, \ldots, \mathbf{v}_T}$ is obtained, where each $\mathbf{v}_t$ captures the spatial characteristics of the corresponding frame. These embeddings are then passed into lightweight adapters for further refinement, as described below:\\

\noindent {\bf Spatial Refinement with Shared Adapter (ShA):}
To adapt frame-level features for DFER, we introduce the Shared Adapter (ShA), a lightweight bottleneck module applied after the MHSA layers of the CLIP visual encoder. The ShA refines spatial feature representations while maintaining computational efficiency through a reduction-expansion mechanism. Following the MHSA operation and layer normalization (LN), the ShA operates as follows:
\begin{equation}
    \mathbf{H}_{\text{MHSA}} = \text{MHSA}(\text{LN}(\mathbf{H}_{\text{input}})).
\end{equation}
The input feature $\mathbf{H}_{\text{MHSA}}$ is reduced using a down-projection matrix $\mathbf{W}_d \in \mathbb{R}^{d/r \times d}$, where $r$ is the reduction factor: 

\begin{equation}
    \mathbf{H}_{\text{down}} = \mathbf{H}_{\text{MHSA}} \cdot \mathbf{W}_d^\top.
\end{equation}
The reduced feature is passed through a GELU activation function to introduce non-linearity: 

\begin{equation}
    \mathbf{H}_{\text{act}} = \text{GELU}(\mathbf{H}_{\text{down}}).
\end{equation}
The original dimensionality is restored using an up-projection matrix $\mathbf{W}u \in \mathbb{R}^{d \times d/r}$: 

\begin{equation}
    \mathbf{H}_{\text{up}} = \mathbf{H}_{\text{act}} \cdot \mathbf{W}_u^\top.
\end{equation}
Finally, the adapter integrates a residual connection by adding its output back to the original input to the MHSA layer: 

\begin{equation}
    \mathbf{H}_{\text{adapt}} = \mathbf{H}_{\text{input}} + \mathbf{H}_{\text{up}}.
\end{equation}
By refining spatial features through this adapter after the MHSA layers, the ShA enables task-specific adaptation without introducing significant computational overhead.\\

\noindent {\bf Temporal Modeling with Temporal Dynamic Adapter:}
Dynamic facial expressions evolve over time, necessitating robust modeling of temporal dependencies. Existing methods, specifically based on the CLIP model, frequently rely on transformer-based temporal models or simple average pooling layers. However, transformer-based approaches, though powerful, are computationally expensive, and average pooling layers often fail to capture subtle temporal dynamics, particularly in short sequences typical of DFER. To address these challenges, we propose the Temporal Dynamic Adapter (TDA), which combines a recurrent neural network, specifically, Gated Recurrent Unit (GRU)~\cite{cho2014learning} for temporal modeling with a learnable dynamic scaling mechanism that adaptively refines temporal feature representations based on their relevance.
The TDA operates after the MLP layers of the visual encoder, 
at its core, the TDA employs a GRU due to its efficiency in modeling sequential dependencies while requiring fewer computational resources compared to transformer-based temporal models. This makes GRU particularly suitable for DFER, where video sequences often comprise 16 frames. The GRU processes a sequence of input features $\mathbf{H}_{\text{adapt}} \in \mathbb{R}^{T \times d}$, where $T$ is the number of frames and $d$ is the feature dimension. To further optimize computational efficiency, the input features are first projected into a lower-dimensional space using a down-projection matrix $\mathbf{W}_d \in \mathbb{R}^{d/r \times d}$, where $r$ is the reduction factor:

\begin{equation}
    \mathbf{H}_{\text{down}} = \mathbf{H}_{\text{MLP}} \cdot \mathbf{W}_d^\top.
\end{equation}
The GRU processes the reduced features $\mathbf{H}_{\text{down}} \in \mathbb{R}^{T \times d/r}$ and computes hidden states for each time step, enabling the model to capture temporal correlations across frames:

\begin{equation}
    \mathbf{h}_t = \text{GRU}(\mathbf{H}_{t,\text{down}}, \mathbf{h}_{t-1}),
\end{equation}\\
where $\mathbf{h}_{t-1}$ is the hidden state from the previous frame. This recurrent mechanism effectively models the subtle dynamics of facial expressions over time, leveraging sequential dependencies in video sequences.\\
To introduce non-linearity and enhance representational capacity, the GRU output $\mathbf{h}_t \in \mathbb{R}^{d/r}$ is passed through a GELU activation function:

\begin{equation}
    \mathbf{h}_t = \text{GELU}(\mathbf{h}_t).
\end{equation}\\
Following this, the activated features are restored to their original dimensionality using an up-projection matrix $\mathbf{W}_u \in \mathbb{R}^{d \times d/r}$:

\begin{equation}
    \mathbf{H}_{t,\text{up}} = \mathbf{h}_t \cdot \mathbf{W}_u^\top.
\end{equation}
By integrating this lightweight yet effective temporal adapter, the TDA strikes a balance between parameter efficiency and the ability to capture temporal dynamics.\\

\noindent {\bf Adaptive Emphasis with Dynamic Scaling:}
In adapter tuning for NLP and 2D vision tasks~\cite{he2021towards, chen2022adaptformer}, the scaling factor plays a pivotal role in determining the importance of features during adaptation. Traditionally, this factor is manually defined, which works well for static tasks but becomes less adequate in dynamic settings like DFER. Video sequences in DFER exhibit varying levels of frame importance; some frames capture critical expression transitions, while others contain less relevant information. A fixed scaling mechanism fails to account for this variability, leading to suboptimal feature adaptation. To address this challenge, we propose a dynamic scaling mechanism inspired by~\cite{zhou2024dynamic}, which enables adaptive weighting of frame's feature representation based on their contribution to temporal modeling.
Instead of assigning uniform importance to all frames, this mechanism learns per-frame scaling factors through a learnable linear transformation, dynamically adjusting the contribution of each frame’s feature representation. This ensures that informative temporal features are emphasized, while noisy or redundant ones are down-weighted, enhancing robustness and efficiency in temporal feature extraction.\\
The dynamic scaling factor is computed from MLP-transformed features before being applied after GRU processing, providing a frame-specific weighting mechanism. Specifically, a learnable linear layer generates scaling factors for each frame's feature representation:

\begin{equation}
    \mathbf{W}_{\text{scale}} = \text{ReLU}\text(linear(\mathbf{H}_{\text{MLP}}))
\end{equation}
where $\mathbf{H}_{t,\text{MLP}} \in \mathbb{R}^d$ represents the MLP-processed feature representation for frame $t$, and $\text{ReLU}$ ensures non-negative scaling factors. These scaling factors are then applied post-GRU to adaptively focus on the most informative temporal variations as follows:

\begin{equation}
    \mathbf{H}_{t,\text{scaled}} = \mathbf{H}_{t,\text{up}} \cdot \mathbf{W}_\text{scale}.
\end{equation}



\subsection{Textual Feature Extraction and Alignment}
\label{sec:ts}
Adapting pre-trained vision-language models such as CLIP for DFER requires constructing semantically rich textual prompts that effectively align with the fine-grained facial expressions, as demonstrated in~\cite{zhao2023prompting},~\cite{li2024cliper}. Existing methods often rely on simple class names or manually crafted prompts (e.g., “A photo of a \{class\}”), which lack the necessary granularity to distinguish subtle expression variations. While advanced prompt learning approaches like CoOp~\cite{zhou2022learning}, CoCoOp~\cite{zhou2022conditional}  introduce learnable tokens to optimize textual prompts, they primarily refine text representations while keeping the visual input unoptimized. This imbalance in adaptation limits their effectiveness for DFER, where subtle temporal and spatial cues contribute to emotion recognition. To address these challenges, we explore AU descriptions as textual prompts, enriching semantic representations with detailed facial muscle movements. Following a similar approach to visual inputs, we incorporate MaPle\_based textual prompts $\mathbf{P}_t$ into enriched textual embeddings. These are then processed through the frozen text encoder, where the ShA adapter refines feature representations, while the TA adapter further enhances the extracted features, ensuring robust and expressive text embeddings for DFER.
\begin{table*}[ht]
\centering
\caption{Facial Expressions and Their Associated Action Units (AUs) Descriptions for FER~\cite{ren2024facial}.}
\label{tab:au}
\resizebox{\textwidth}{!}{ 
\begin{tabular}{lccc}
\toprule
\textbf{Facial Basic Expressions}  & \textbf{AU-Based Descriptions}\\ 
\midrule

Happiness& Cheek Raiser, Lip Corner Puller. \\ 
Sadness & Inner Brow Raiser, Brow Lowerer, Lip Corner Depressor  \\ 
Neutral& Relaxed Muscles, Even Eyebrows, Closed Lips, Calm Eyes, Smooth Forehead \\ 
Anger & Brow Lowerer, Upper Lid Raiser, Lid Tightener, Lip Tightener \\ 
Surprise &Inner Brow Raiser, Outer Brow Raiser, Upper Lid Raiser, Jaw Drop  \\ 
Disgust& Nose Wrinkler, Lip Corner Depressor, Lower Lip Depressor \\ 
Fear&Inner Brow Raiser, Outer Brow Raiser, Brow Lowerer, Upper Lid Raiser, Lid Tightener, Lip Stretcher, Jaw Drop \\ 

\bottomrule
\end{tabular}
}
\end{table*}

\noindent {\bf Leveraging Action Units for Enriched Textual Prompts:}
Traditional textual prompts in DFER often lack semantic specificity, as they rely on abstract class names such as "Surprise" or "Fear." These simplistic prompts fail to capture detailed contextual information about the facial features and muscle movements that define each expression, thereby limiting their effectiveness in aligning with visual features. Inspired by works such as~\cite{du2014compound, ren2024facial}, which emphasize the relationship between facial expressions and AUs defined in the facial action coding system, we incorporate AU-based descriptions into textual prompts to enrich their semantic content. AUs describe specific facial muscle movements, providing a granular breakdown of facial dynamics. For example, as shown in Table~\ref{tab:au}, the expression "Surprise" is associated with features such as Inner Brow Raiser (AU1), Outer Brow Raiser (AU2), and Jaw Drop (AU26). By leveraging these AU-based descriptions, we provide semantically rich prompts that better capture the nuances of facial expressions, allowing for stronger alignment with expression-relevant visual features. Formally, for a given expression class \(c_k\), the textual prompt \(T_k\) is directly constructed using the fine-grained AUs descriptions associated with the class and represented as:

\[
T_k = \{\text{AU}_1, \text{AU}_2, \dots, \text{AU}_n\},
\]
where \(\text{AU}_i\) denotes the textual description of the \(i\)-th Action Unit corresponding to the expression. These fine-grained prompts replace the abstract class names with semantically rich descriptions of facial muscle movements, enabling the model to better align visual features with textual information.\\


\noindent {\bf Enhancing Alignment with Multi-modal Prompt Learning (MaPLe):} 
To improve the alignment between visual and textual modalities, we integrate Multi-modal Prompt Learning, a framework designed to jointly adapt both image and text encoders in vision-language models~\cite{khattak2023maple}. Unlike single-modality prompt learning approaches, MaPLe introduces learnable tokens into both the textual and visual inputs, improving alignment. In MaPLe, the textual encoder processes AU-enriched textual prompts $\mathbf{T}_k$ (as detailed earlier) along with learnable tokens $\mathbf{P}_t \in \mathbb{R}^{N_t \times d}$, where $N_t$ is the number of learnable tokens, and $d$ is the embedding dimension. These learnable tokens, incorporated with the AU-Tokenized text to form $\mathbf{C}_t$, which are then passed to the textual encoder to produce the refined textual representation:
\begin{equation}
    \mathbf{\textit{f}}_{t} = \text{TextEncoder}(\mathbf{C}_t),
\end{equation}
where $\mathbf{C}_t$ are textual embeddings. Similarly, MaPLe adapts the visual encoder by introducing learnable tokens $\mathbf{P}_v \in \mathbb{R}^{N_v \times d}$, where $N_v$ is the number of visual learnable tokens. These tokens are prepended to the visual embedding to form $\mathbf{X}_v$, allowing the vision encoder to refine its representations and better align with the textual embeddings. The refined visual representation $\mathbf{\textit{f}}_{v}$ is computed as:
\begin{equation}
    \mathbf{\textit{f}}_{v} = \text{ImageEncoder}(\mathbf{X}_v),
\end{equation}
To further enhance this alignment, MaPLe conditions the vision prompts on language prompts via a vision-language (V-L) coupling function, inducing mutual synergy between the two modalities.
\subsection{Classification via Cosine Similarity}  
\label{sec:class}

Leveraging the shared embedding space of CLIP, the refined visual representation $\mathbf{\textit{f}}_{v}$, which encapsulates the temporal features of the video sequence, is aligned with the refined textual embeddings $\mathbf{\textit{f}}_{t}$. The cosine similarity metric quantifies the alignment between these embeddings, ensuring effective matching of the temporal with the semantic information embedded in the textual prompts (See Equation~\eqref{eq:cosine_similarity}). Since the cosine similarity scores do not inherently represent probabilities, we optimize the model using a cross-entropy loss function, which internally applies a softmax operation to normalize the scores. This ensures that the model learns a probability distribution over the classes, making the classification process robust. 
Finally, the emotion class with the highest probability is selected as the predicted class.

\section{Experimental Analysis}
\label{sec:exp}

To assess the performance of PE-CLIP, we performed extensive experiments on two widely-used benchmark datasets: DFEW~\cite{jiang2020dfew} and FERV39k~\cite{wang2022ferv39k}. A brief overview of these datasets is provided in Section~\ref{sec:data}. Section~\ref{sec:impl} outlines the implementation setup. The results and comparison against state-of-the-art methods on both datasets are presented in Section~\ref{sec:exp2}. Furthermore, Section~\ref{sec:ab} includes ablation studies conducted on the DFEW dataset to analyze the contributions of each individual component of our framework. Finally, Section~\ref{sec:vis} presents qualitative visualizations that illustrate the model's behavior.

\subsection{Dataset Description}
\label{sec:data}

\noindent {\bf DFEW}~\cite{jiang2020dfew} is a large-scale, in-the-wild benchmark dataset introduced in 2020. It is one of the most widely used benchmarks for dynamic facial expression recognition, consisting of over 16,000 video clips extracted from more than 1,500 movies worldwide. The dataset encompasses several challenging scenarios, including extreme illumination conditions, occlusions, and variations in head poses. Each video clip is labeled by professional annotators ten times and classified into one of seven basic facial expressions: happiness, sadness, neutral, anger, surprise, disgust, and fear. The dataset is divided into five equal, non-overlapping subsets, and a 5-fold cross-validation protocol is adopted to ensure consistency and fair evaluation.\\

\noindent {\bf FERV39K} ~\cite{wang2022ferv39k}, released in 2022, is currently the largest publicly available in-the-wild DFER dataset. It comprises 38,935 video clips collected across four scenarios and further categorized into 22 fine-grained scenes. To ensure high-quality annotations, each clip is labeled by 30 professional annotators and classified into one of the seven basic expressions, similar to DFEW. The dataset is randomly divided into a training set (31,088 clips) and a testing set (7,847 clips), corresponding to 80\% and 20\% of the data, respectively. For consistency and fair comparison, we utilized the officially provided training and testing splits.

\subsection{Implementation Setup}
\label{sec:impl}
\noindent {\bf Data Pre-processing:}
For both DFEW and FERV39K datasets, the preprocessed facial regions of video frames, as provided in the datasets, are directly utilized.  Each video sequence is preprocessed by resizing and cropping 16 consecutive frames to a uniform size of $224 \times 224$ pixels. To mitigate overfitting and enhance generalization, various data augmentation techniques are applied, including random cropping, random horizontal flipping, and color jittering, as suggested in~\cite{Zhao_2023_BMVC, sun2023mae, chen2024static}.  Additionally, to further increase the diversity of training data, we integrate Mixup~\cite{zhang2017mixup} and FMix~\cite{harris2020fmix} techniques. During training, a random probability \( p \) determines the type of augmentation applied. If \( p < 0.4 \), Mixup is used, blending data and targets using a Beta distribution with a parameter \( \alpha=0.4 \). If \( p \) is between 0.4 and 0.7, FMix generates binary masks based on the given \( \alpha\) and decay power, combining the input data with shuffled samples to introduce new variations. For \( p \geq 0.7 \), no augmentation is applied. This dynamic augmentation strategy balances the original and augmented samples, enhancing the model's robustness to variations in training data and improving performance.

\vspace{2mm}
\noindent {\bf Training Setting:}
All experiments for the proposed PE-CLIP framework are conducted using the CLIP model with a ViT-B/16 backbone, implemented in PyTorch. Training is performed on two NVIDIA Quadro RTX 5000 GPUs, each with 16 GB of memory. To ensure parameter efficiency, only the adapter and prompt learner parameters, totaling 9M trainable parameters, are fine-tuned, while the visual and text encoders remain frozen. The adapters employ a reduction factor \( r \) of 8 to achieve parameter efficiency without sacrificing performance. 
For the DFEW dataset, training is performed for 30 epochs in an end-to-end manner. The AdamW optimizer is used with an initial learning rate of $2 \times 10^{-5}$, adjusted via a cosine learning rate scheduler with a warmup period of 3 epochs. Separate learning rates are set for different components: $2 \times 10^{-4}$ for the adapters and $3 \times 10^{-5}$ for the prompt tokens. The weight decay is set to 0.01, and the batch size is set to 8. For the FERV39K dataset, the model is trained for 20 epochs using the AdamW optimizer and cosine learning rate scheduler with a warmup period of 2 epochs. The initial learning rate is $4 \times 10^{-5}$, while the weight decay and batch size remain consistent with the DFEW settings. These configurations used in our experiments are illustrated in Table~\ref{tab:hyperparameters}.

\begin{table}[ht]
\centering
\caption{Hyperparameter Configurations for Training PE-CLIP on DFEW and FERV39K Datasets.}
\label{tab:hyperparameters}
\begin{tabular}{ccc}
\toprule
\textbf{Hyperparameters} & \textbf{DFEW} & \textbf{FERV39K} \\ 
\midrule
Batch size             & 8   & 8             \\ 
Learning rate (adapters) & $2 \times 10^{-4}$ & $2 \times 10^{-4}$\\ 
Learning rate (prompts)  & $3 \times 10^{-5}$ & $3 \times 10^{-5}$\\ 
Weight decay            & 0.01    & 0.01          \\
Optimizer               & AdamW  & AdamW          \\ 
Scheduler               & CosineLR & CosineLR \\ 
Warmup                  & 3 & 2 \\ 
Reduction factor        & 8  & 8               \\ 
Training epochs         & 30  & 20             \\ 
\bottomrule
\end{tabular}
\end{table}

\noindent {\bf Evaluation Metrics:}
We evaluate the performance of our model using two key metrics: Weighted Average Recall (WAR) and Unweighted Average Recall (UAR). WAR accounts for class distribution by weighting each class's recall according to its frequency, making it particularly relevant for imbalanced datasets. UAR, in contrast, treats all classes equally, providing a balanced evaluation of performance across categories. Consistent with prior studies, we report both metrics.


\subsection{Experimental Results}
\label{sec:exp2}
We evaluated the effectiveness of our proposed PE-CLIP method, by comparing it against existing state-of-the-art methods on two benchmark in-the-wild datasets, DFEW and FERV39K. 

\noindent {\bf Results on DFEW:}
Table~\ref{tab:res1} presents the average performance of PE-CLIP across 5-fold cross-validation on the DFEW dataset. Our proposed method achieves a UAR of 62.82\% and a WAR of 74.04\%, demonstrating its effectiveness against state-of-the-art methods. Notably, PE-CLIP outperforms vision-only approaches such as M3DFEL\cite{wang2023rethinking} and Dual-STI\cite{li2024dual}, and achieves an improvement of 6.16\% in UAR and 4.67\% in WAR over AEN~\cite{lee2023frame}. This highlights its ability to extract richer, more discriminative features for dynamic facial expression modeling. When compared to CLIP-based methods, PE-CLIP consistently surpasses their performance, with improvements of 3.21\% in UAR and 2.79\% in WAR over DFER-CLIP. Furthermore, PE-CLIP outperforms the recently proposed CLIP-Guided-DFER~\cite{10744485}, achieving a UAR gain of 1.97\% and a WAR gain of 1.46\%, underscoring its ability to 
better align the vision and language representations through the adapters and prompt learning.
\begin{table*}[htpb]
\centering
\caption{Comparison of PE-CLIP with state-of-the-art methods on the DFEW and FERV39K datasets. The table highlights the Unweighted Average Recall (UAR) and Weighted Average Recall (WAR) achieved by each method, alongside their respective backbones and the number of tunable parameters (in millions).}
\label{tab:res1}
\begin{tabular}{l c c c c c c c}
\hline
\textbf{Method} & \textbf{Backbone} & \textbf{Tunable Param (M)} & \multicolumn{2}{c}{\textbf{DFEW}} & \multicolumn{2}{c}{\textbf{FERV39k}} \\
\cline{4-7} 
 & & & \textbf{UAR} & \textbf{WAR} & \textbf{UAR} & \textbf{WAR} \\
\hline
EC-STFL (MM'20) \cite{jiang2020dfew} & C3D / P3D & 78 & 45.35 & 56.51 & - & - \\
Former-DFER (MM'21) \cite{zhao2021former} & Transformer & 18 & 53.69 & 65.70 & 37.20 & 46.85 \\
CEFLNet (IS'22) \cite{liu2022clip} & ResNet-18 & 13 & 51.14 & 65.35 & - & - \\
NR-DFERNet (ArXiv'22) \cite{li2022nr} & CNN-Transformer & - & 54.21 & 68.19 & 33.99 & 45.97 \\
STT (ArXiv'22) \cite{ma2022spatio} & ResNet-18 & - & 54.58 & 66.65 & 37.76 & 48.11 \\
DPCNet (MM'22) \cite{wang2022dpcnet} & ResNet-50 & - & 57.11 & 66.32 & - & -\\
EST (PR'23) \cite{liu2023expression} & ResNet-18 & 43 & 53.94 & 65.85 & - & - \\
Freq-HD (MM'23) \cite{tao2023freq} & VGG13-LSTM & - & 46.85 & 55.68 & 33.07 & 45.26 \\
LOGO-Former (ICASSP'23) \cite{ma2023logo} & ResNet-18 & - & 54.21 & 66.98 & 38.22 & 48.13 \\
IAL (AAAI'23) \cite{li2023intensity} & ResNet-18 & 19 & 55.71 & 69.24 & 35.82 & 48.54 \\
AEN (CVPRW'23) \cite{lee2023frame} & ResNet-18 & - & 56.66 & 69.37 & 38.18 & 47.88 \\
M3DFEL (CVPR'23) \cite{wang2023rethinking} & ResNet-18-3D & - & 56.10 & 69.25 & 35.94 & 47.67 \\
Dual-STI (IS'24) \cite{li2024dual} & ResNet-18 & - & 56.89 & 68.29 & 38.27  & 48.46 \\
DFER-CLIP (BMVC'23) \cite{zhao2023prompting} & CLIP-ViT-B/32 & 90 & 59.61& 71.25 & 41.27 & 51.65 \\
CLIPER (ICME'24) \cite{li2024cliper} & CLIP-ViT-B/16 & 88 & 57.56 & 70.84 & 41.23 & 51.34 \\
EmoCLIP (FG'24) \cite{foteinopoulou2024emoclip} & CLIP-ViT-B/32 & - & 58.04 & 62.12 & 31.41 & 36.18 \\
DFLM (ICSP'24) \cite{10743836} & CLIP-ViT-B/32 & - & 59.77 & 71.40 & 41.25 & 51.31 \\
CLIP-Guided-DFER (IJCB'24) \cite{10744485} & CLIP-ViT-B/32 & - & \underline{60.85} & \underline{72.58} & \underline{41.43} & \textbf{51.83} \\
\hline
\rowcolor{green!10}\textbf{PE-CLIP (ours)}& CLIP-ViT-B/16 & \textbf{9} & \textbf{62.82} & \textbf{74.04} & \textbf{41.57} & \underline{51.26} \\
\hline
\end{tabular}
\end{table*}

Furthermore, as shown in Table~\ref{tab:res2}, PE-CLIP demonstrates remarkable performance across various emotional categories on the DFEW dataset, further validating its robustness and effectiveness. Specifically, our method achieves 91.00\% average accuracy on "happiness," the highest among all methods, benefiting from the distinct visual features and high representation of this emotion in the dataset. Similarly, for "sadness," "neutral," "angry," and "surprise," PE-CLIP consistently achieves high accuracy, demonstrating its ability to capture visually distinctive and high-intensity expressions. In more challenging categories, such as "fear" and "disgust," where data imbalance poses a significant challenge, PE-CLIP achieves superior results. For "fear," our model attains 45.12\% average accuracy, marking a notable 8.68\% improvement over the previous best method, IAL~\cite{li2023intensity}, reflecting its ability to capture nuanced temporal features in less prominent emotions. Similarly, for the underrepresented "disgust," PE-CLIP reaches 15.17\% average accuracy, achieving an 6.20\% improvement over Dual-STI~\cite{li2024dual}, further highlighting its effectiveness in handling both well-represented and challenging emotional categories.

  
\noindent {\bf Results on FERV39K:}
The performance comparison of PE-CLIP with existing methods on the FERV39K dataset is presented in Table~\ref{tab:res1}. Our approach achieves a UAR of 41.57\% and a WAR of 51.26\%, demonstrating superior results compared to vision-only based methods, and competitive results, particularly against CLIP-based methods. Compared to CLIP-Guided-DFER, PE-CLIP shows a marginal UAR improvement of +0.14\%, while WAR exhibits a slight decline of -0.57\%, which may be attributed to the increased variability and complexity of emotional representations in FERV39K. Unlike DFEW, which features more clearly distinguishable expressions, FERV39K introduces fine-grained nuances and imbalanced class distributions that pose challenges for all existing methods. Despite these factors, PE-CLIP maintains competitive performance, demonstrating its robustness in handling dynamic expressions across different datasets.

\begin{table*}[h!]
\centering
\caption{Accuracy Comparison Among Various Emotion Categories on DFEW.}
\label{tab:res2}
\resizebox{\textwidth}{!}{ 
\begin{tabular}{lcccccccccccc}
\hline
\textbf{Method} & \textbf{Tunable Param (M)} & \multicolumn{7}{c}{\textbf{Accuracy of Each Emotion}} & \multicolumn{2}{c}{\textbf{DFEW}} \\
\cline{3-11} 
 & & \textbf{Hap.} & \textbf{Sad.} & \textbf{Neu.} & \textbf{Ang.} & \textbf{Sur.} & \textbf{Dis.} & \textbf{Fea.} & \textbf{UAR} & \textbf{WAR} \\
\hline

Former-DFER (MM'21) \cite{zhao2021former} & 18 & 84.05 & 62.57 & 67.52 & 70.03 & 56.43 & 3.45 & 31.78 & 53.69 & 65.70 \\
CEFLNet (IS'22) \cite{liu2022clip} & 13 & 84.00 & 68.00 & 67.00 & 70.00 & 52.00 & 0.00 & 17.00 & 51.14 & 65.35 \\
NR-DFERNet (ArXiv'22) \cite{li2022nr} & - & 88.47 & 64.84 & 70.03 & 75.09 & 61.60 & 0.00 & 19.43 & 54.21 & 68.19 \\
STT (ArXiv'22) \cite{ma2022spatio} & - & 87.36 & 67.90 & 64.97 & 71.24 & 53.10 & 3.49 & 34.04 & 54.58 & 66.65 \\
EST (PR'23) \cite{liu2023expression} & 43 & 86.87 & 66.58 & 67.18 & 71.84 & 47.53 & 5.52 & 28.49 & 53.43 & 65.85 \\
IAL (AAAI'23)  \cite{li2023intensity} & 19 & 87.95 & 67.21 & \underline{70.10} & \underline{76.06} & \underline{62.22} & 0.00 & \underline{36.44} & 55.71 & 69.24 \\
M3DFEL (CVPR'23) \cite{wang2023rethinking} & - & \underline{89.59} & \underline{68.38} & 67.88 & 74.24 & 59.69 & 0.00 & 31.64 & \underline{56.10} & \underline{69.25} \\
Dual-STI (IS'24) \cite{li2024dual}  & - & 87.65 & 68.32 & 67.55 & 71.70 & 57.93 & \underline{8.97} & 36.14 & 55.71 & 69.24 \\    
\hline
\rowcolor{green!10}PE-CLIP (ours) & \textbf{9} & \textbf{91.00} & \textbf{79.62} & \textbf{71.94} & \textbf{79.80} & \textbf{85.10} & \textbf{15.17} & \textbf{45.12} & \textbf{62.82} & \textbf{74.04} \\
\hline
\end{tabular}
}
\end{table*}

\noindent {\bf About Model Size Efficiency:}
While PE-CLIP achieves competitive performance, its efficiency remains a key distinguishing factor. As shown in Table~\ref{tab:res1}, our approach requires only 9M trainable parameters, significantly fewer than most existing methods while maintaining high performance. Compared to vision-only methods, such as Former-DFER (18M), CEFLNet (13M), and IAL (19M), PE-CLIP reduces the trainable parameter count by nearly 50\% while achieving superior results. This demonstrates the advantage of using vision-language alignment for dynamic expression recognition over purely vision-based approaches. Among CLIP-based methods, such as DFER-CLIP (90M) and CLIPER (88M), which require nearly 10× more trainable parameters, PE-CLIP offers a substantial reduction in trainable parameters, achieving an impressive 90\% reduction. In summary, we can conclude that PE-CLIP achieves an optimal balance between parameter efficiency and performance, requiring fewer than 6\% of the model's parameters to be trained while avoiding full fine-tuning. These results highlight the effectiveness of PE-CLIP’s parameter-efficient design and its practicality for DFER.
\subsection{Ablation Analysis}
\label{sec:ab}

To gain insights into the effectiveness of the key components in PE-CLIP, we conducted a series of ablation studies on the DFEW dataset. The analysis begun with an assessment of the overall architecture. We then examined the contributions of the main components, including the proposed vision and textual adapters. Following this, we analyzed the impact of various temporal modeling techniques. Additionally, we analyzed how different reduction factor settings influence model performance and computational efficiency in terms of trainable parameters. Finally, we investigated the effect of textual prompting methods on alignment and overall performance. For consistency and clarity, all results are reported for fold 1 (fd1), aligning with prior studies~\cite{sun2023mae,chen2024static,chen2024finecliper}.

\noindent {\bf Evaluation of the Overall Framework:}
Table~\ref{tab:t1} provides a detailed comparison of the performance of PE-CLIP overall framework and its key general components on the DFEW dataset. The baseline model, comprising frozen CLIP encoders with a single trainable vanilla adapter \cite{houlsby2019parameter} integrated into the textual encoder after the MLP layers and class names as textual inputs, achieves a UAR of 49.82\% and a WAR of 63.35\% with only 0.8M trainable parameters. While effective in leveraging CLIP’s pre-trained capabilities for generalizing to DFER, the baseline's performance is limited by its inability to fully capture temporal information and multimodal complexities. Adding lightweight adapters to both the vision and text encoders significantly enhances performance, achieving a UAR of 61.74\% and a WAR of 74.64\%. This improvement of 11.92\% in UAR and 11.29\% in WAR underscores the effectiveness of adapters in refining pre-trained features for DFER while maintaining computational efficiency. Finally, incorporating Multi-modal Prompt Learning (MaPLe) with AU-based descriptions into the framework yields further gains, with an improvement of 1.28\% in UAR and 0.71\% in WAR. This progression highlights the critical role of semantically rich textual prompts in enhancing vision-language alignment. These results validate the overall framework of PE-CLIP, emphasizing the important  contributions of the adapters and prompt learning in achieving competitive performance on DFER with only 9M trainable parameters.

\begin{table}[h!]
\centering
\caption{Performance Comparison of the Overall Framework Components on DFEW Dataset.}
\label{tab:t1}
\begin{tabular}{lccc}
\toprule
\textbf{Model}  & \textbf{Params (M)} & \multicolumn{2}{c}{\textbf{DFEW}} \\ 
\cmidrule(lr){3-4}
& & \textbf{UAR} & \textbf{WAR} \\ 
\midrule

Baseline & 0.8 & 49.82 & 63.35 \\ 
Baseline + Adapters &6& 61.74 & 74.64 \\ 
\rowcolor{green!10}Baseline + Adapters + Prompts (Ours)&9&\textbf{63.02} & \textbf{75.35} \\ 
\bottomrule
\end{tabular}
\end{table}

\vspace{2mm}
\noindent {\bf Impact of the Proposed Adapters in PE-CLIP:} 
Table~\ref{tab:tdap} shows the impact of different adapter configurations within PE-CLIP, demonstrating their individual and combined contributions to DFER. All experiments in this analysis incorporate AU-MaPLe prompts to maintain a consistent textual prompting strategy. In the first configuration, with a single Textual Adapter (TA) placed after the MLP layers, the model achieves a UAR of 53.64\% and a WAR of 68.18\%, utilizing 4M trainable parameters. This relatively lower performance highlights the limitations of relying solely on textual adaptation, as it inadequately aligns textual features with the visual features essential for DFER. Introducing a second textual adapter after the MHSA layers leads to a UAR increase to 55.43\% (+1.79\%), and a slight WAR decrease to 68.09\% (-0.09\%). This trade-off could be attributed to a disparity in how textual features align with the dataset’s class distribution, where textual features improve recall across classes but lack complementary vision adapters, reducing the model's ability to generalize effectively to frequent emotions.
Shifting the focus to vision adapters, a configuration incorporating a Shared Adapter (ShA) after the MHSA layers and a Temporal Dynamic Adapter (TDA) after the MLP layers yields substantial performance gains, achieving a UAR of 62.59\% and a WAR of 74.43\% with 8M trainable parameters. This represents a +7.16\% UAR and +6.34\% WAR improvement compared to the dual-text-adapter configuration, emphasizing the critical role of vision adapters in refining spatial features and capturing temporal dependencies. particularly through TDA’s dynamic scaling mechanism, which prioritizes the most informative temporal features. However, the lack of semantic enrichment from textual embeddings limits the model’s ability to capture finer-grained expression nuances. Introducing ShA across both text and visual encoders while retaining the TA adapter results in a UAR of 60.76\% and a WAR of 72.70\%, utilizing 7M trainable parameters. This configuration benefits from ShA’s ability to effectively refine the extracted features in both encoders. However, the absence of TDA restricts its capability to fully capture temporal variations, leading to slightly lower performance than the previous configuration. Finally, integrating all adapters, TA, ShA, and TDA, achieves the best performance, with a UAR of 63.02\% and a WAR of 75.35\%, using 9M trainable parameters. Notably, adding TDA leads to a +2.26\% UAR and +2.65\% WAR increase compared to the configuration without it, highlighting the important role of the temporal modeling in DFER, where effectively capturing dynamic relationships across frames is crucial. These findings confirm the complementary nature of each adapter in PE-CLIP framework, allowing it to achieve good performance while maintaining efficiency with just 9M trainable parameters.

\begin{table}[ht]
\centering
\caption{Impact of Different Adapters on PE-CLIP Performance on the DFEW Dataset (TA: Textual Adapter, ShA: Shared Adapter, TDA: Temporal Dynamic Adapter).}
\label{tab:tdap}
\begin{tabular}{ccccccc}
\toprule
\multicolumn{2}{c}{\textbf{Text}} & \multicolumn{2}{c}{\textbf{Video}} & \textbf{Params (M)} & \multicolumn{2}{c}{\textbf{DFEW}} \\
\cmidrule(lr){1-2} \cmidrule(lr){3-4} \cmidrule(lr){6-7}
TA & ShA & ShA & TDA & & UAR & WAR \\
\midrule
\checkmark & \xmark & \xmark & \xmark & 4 & 53.64 & 68.18 \\
\checkmark & \checkmark & \xmark & \xmark & 5 & 55.43 & 68.09 \\
\xmark & \xmark & \checkmark & \checkmark & 8 & 62.59 & 74.43 \\
\checkmark & \checkmark & \checkmark & \xmark & 7 & 60.76 & 72.70 \\
\rowcolor{green!10}\checkmark & \checkmark & \checkmark & \checkmark & 9 & \textbf{63.02} & \textbf{75.35} \\
\bottomrule
\end{tabular}
\end{table}
\vspace{2mm}
\noindent {\bf Impact of Dynamic Scaling in Temporal Dynamic Adapter (TDA):} Table~\ref{tab:t3} evaluates the influence of dynamic scaling applied at different stages within the TDA adapter, demonstrating its importance in enhancing temporal modeling. Without dynamic scaling, the TDA achieves a UAR of 62.10\% and a WAR of 75.22\%. While the GRU effectively captures temporal dependencies, the absence of frame-specific weighting limits the model’s ability to adaptively emphasize salient temporal features, establishing this configuration as a baseline. Applying dynamic scaling at the input level before GRU processing improves UAR by 0.63\% but slightly reduces WAR by 0.94\%. The increase in UAR highlights an improved focus on underrepresented classes, as input-level scaling helps prioritize discriminative temporal patterns before they are modeled. However, the slight decrease in WAR suggests that early-stage scaling may disrupt the natural temporal coherence modeled by the GRU, leading to overemphasis on certain temporal variations. When applied directly to the GRU’s output, dynamic scaling results in decreases of 0.48\% in UAR and 0.55\% in WAR, likely due to interference with the refined temporal dependencies captured by the GRU. Weighting at this stage alters the consistency of learned representations, reducing the effectiveness of the adaptation. The best performance is achieved when dynamic scaling is applied after the up-projection layer, yielding a UAR of 63.02\% (+0.92\%) and a WAR of 75.35\% (+0.13\%) compared to the baseline. This approach ensures adaptive weighting of the refined temporal features after dependencies are fully modeled, while effectively enhancing salient temporal cues and mitigating irrelevant or noisy variations. Across all configurations, the number of trainable parameters remains constant at 9M, as dynamic scaling introduces only a negligible computational overhead. These results highlight the efficiency and effectiveness of incorporating dynamic scaling into the TDA, enabling the model to achieve optimal performance without compromising its lightweight design.

\begin{table}[ht]
\centering
\caption{Impact of dynamic scale mechanism in TDA. (w/o: without, DS: dynamic Scale ).}
\label{tab:t3}
\begin{tabular}{lccc}
\toprule
\textbf{Setting}  & \textbf{Params (M)} & \multicolumn{2}{c}{\textbf{DFEW}} \\ 
\cmidrule(lr){3-4}
& & \textbf{UAR} & \textbf{WAR} \\ 
\midrule

TDA w/o DS& 9 &62.10 & 75.22 \\ 
TDA w/ Input-Level Scaling&9& 62.73 & 74.28 \\ 
TDA w/ GRU Output Scaling&9& 61.62 & 74.67 \\ 
\rowcolor{green!10}TDA w/ Post-Up-Projection Scaling (Ours)&9& \textbf{63.02} & \textbf{75.35} \\

\bottomrule
\end{tabular}
\end{table}
\vspace{2mm}
\noindent {\bf Impact of Temporal Modeling Architectures on PE-CLIP Performance:}
We evaluated the performance of PE-CLIP using different RNN-based temporal adapter configurations within TDA, as detailed in Table~\ref{tab:t4}. The Vanilla Adapter, which lacks temporal modeling, achieves a UAR of 58.80\% and a WAR of 72.23\%, demonstrating its inability to capture sequential dependencies across frames. Introducing an RNN-based temporal adapter improves UAR to 62.18\% (+3.38\%) and WAR to 73.86\% (+1.63\%), demonstrating enhanced temporal modeling. However, its limited ability to balance short- and long-term dependencies restricts its effectiveness. The LSTM-based adapter, an advanced RNN variant with gated memory mechanisms, further improves WAR to 74.16\% (+1.93\%) but slightly reduces UAR to 61.29\% (-0.89\% compared to the RNN-based adapter). This suggests that LSTM’s long-term dependency modeling benefits dominant classes while being slightly less sensitive to underrepresented ones. The GRU-based adapter, a computationally efficient alternative to LSTM, achieves the best performance, with a UAR of 63.02\% (+4.22\%) and WAR of 75.35\% (+3.12\%) compared to the baseline, outperforming the LSTM-based adapter by +1.73\% in UAR and +1.19\% in WAR. These results highlight GRU’s efficiency in balancing computational cost and temporal accuracy due to its gated recurrent mechanism, making it particularly suitable for DFER.

\begin{table}[ht]
\centering
\caption{Performance Comparison of Temporal Adapter Configurations within TDA on DFEW.}
\label{tab:t4}
\begin{tabular}{lccc}
\toprule
\textbf{Temporal Adapter}  & \textbf{Params (M)} & \multicolumn{2}{c}{\textbf{DFEW}} \\ 
\cmidrule(lr){3-4}
& & \textbf{UAR} & \textbf{WAR} \\ 
\midrule

Vanilla Adapter\cite{houlsby2019parameter}&9& 58.80 & 72.23 \\ 
RNN-based Adapter&9& 62.18 & 73.86 \\ 
LSTM-based Adapter&10&61.29& 74.16 \\ 
\rowcolor{green!10}GRU-based Adapter (Ours)&9& \textbf{63.02} & \textbf{75.35} \\ 

\bottomrule
\end{tabular}
\end{table}

\vspace{3mm}
\noindent {\bf Effect of Reduction Factor on Efficiency and Model Performance:}
Table~\ref{tab:t5} shows  the impact of varying reduction factors on the efficiency and performance of PE-CLIP. The reduction factor determines the degree of dimensionality reduction within the bottleneck layers of adapters, directly affecting both trainable parameters and recognition accuracy. With a reduction factor of 4, the model utilizes 17M trainable parameters, achieving a UAR of 61.39\% and a WAR of 73.86\%. While this setting effectively captures features, its higher parameter count reduces practicality for resource-constrained scenarios. Increasing the reduction factor to 8 reduces trainable parameters to 9M, while improving performance to a UAR of 63.02\% (+1.63\%) and a WAR of 75.35\% (+1.49\%) compared to the reduction factor of 4. This configuration strikes the best balance between dimensionality reduction and representation quality, demonstrating its suitability for DFER. Further increasing of the reduction factor to 16 lowers the trainable parameters to 6M but slightly decreases performance, with a UAR of 62.68\% (-0.34\%) and a WAR of 74.75\% (-0.60\%). This suggests that too much reduction constrains the adapters’ ability to retain important temporal and spatial features, leading to performance degradation. Overall, a reduction factor of 8 offers the optimal trade-off, achieving a highly parameter-efficient design while maintaining robust performance, making it well-suited for DFER.

\begin{table}[ht]
\centering
\caption{Influence of reduction factor of adapters.}
\label{tab:t5}
\begin{tabular}{lccc}
\toprule
\textbf{Reduction Factor}  & \textbf{Params (M)} & \multicolumn{2}{c}{\textbf{DFEW}} \\ 
\cmidrule(lr){3-4}
& & \textbf{UAR} & \textbf{WAR} \\ 
\midrule

4& 17 &61.39 & 73.86 \\ 
\rowcolor{green!10}8 (Ours)&9& \textbf{63.02} & \textbf{75.35} \\ 
16&6& 62.68 & 74.75 \\

\bottomrule
\end{tabular}
\end{table}
\noindent {\bf Impact of Textual Prompt:}
We analyzed the impact of different textual inputs and prompting mechanisms on the PE-CLIP's performance, as shown in Table~\ref{tab:t6}. Using only class names (e.g., "Happiness", "Anger") the model achieves a UAR of 61.74 \% and a WAR of 74.46\%/. However, the lack of semantic richness limits the model’s ability to capture nuanced expressions. Introducing learnable prompts via CoOp~\cite{zhou2022learning}, which incorporates a generic learnable token into textual inputs, results in a performance decline of (-0.82\%) UAR and (-1.54\%) WAR compared to class names, highlighting its lack of task-specific guidance. Replacing class names with AU-based descriptions within the CoOp framework improves performance, yielding a +0.31\% UAR and +0.81\% WAR increase, demonstrating the value of richer semantic information from AUs. However, CoOp’s generic token learning still restricts its effectiveness. A task-specific prompting approach using ChatGPT-generated descriptions~\cite{zhao2023prompting} further enhances UAR by +1.63\% and WAR by +0.21\%, underscoring the importance of tailored prompts for textual-visual alignment. Finally, MaPLe with AU-based descriptions, which introduces learnable tokens for both textual and visual inputs, achieves the highest WAR (75.35\%) and a UAR of (63.02\%). Compared to CoOp with AU descriptions, MaPLe demonstrates a substantial improvement of +1.79\% UAR and +1.62\% WAR, highlighting the effectiveness of multi-modal prompting in dynamic facial expression recognition. However, compared to ChatGPT-based prompting~\cite{zhao2023prompting}, MaPLe achieves a slightly lower UAR (-0.35\%) but a higher WAR (+0.68\%), suggesting that ChatGPT-based prompts better support underrepresented classes, whereas MaPLe with AU descriptions optimizes alignment for dominant classes, enhancing overall performance.

\begin{table}[ht]
\centering
\caption{Performance comparison of key components on the DFEW dataset.}
\label{tab:t6}
\begin{tabular}{lcc}
\toprule
\textbf{Setting} & \multicolumn{2}{c}{\textbf{DFEW}} \\ 
\cmidrule(lr){2-3}
 & \textbf{UAR} & \textbf{WAR} \\ 
\midrule

Class Names & 61.74 & 74.46 \\
CoOp (Generic Learnable Token)~\cite{zhou2022learning} & 60.92 & 72.92 \\ 
CoOp + AU Descriptions & 61.23 & 73.73\\ 
Learnable prompt by ChatGPT~\cite{Zhao_2023_BMVC}& \textbf{63.37} & 74.67 \\ 

\rowcolor{green!10}MaPLe + AU Descriptions (Ours) &  \underline{63.02} & \textbf{75.35} \\ 
\bottomrule
\end{tabular}
\end{table}

\subsection{Visualization}
\label{sec:vis}

\begin{figure*}[!h]
      \centering
      \includegraphics[width=1.0\linewidth]{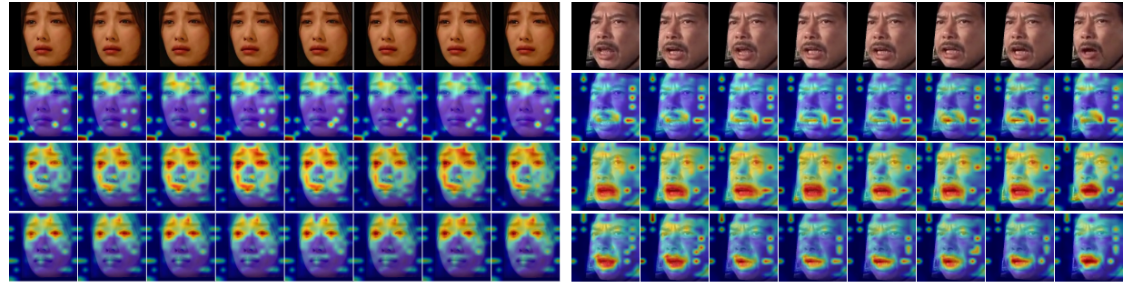}

      \caption{Attention visualization of our proposed PE-CLIP model. The figure presents attention maps for \textit{sadness} and \textit{anger} (left to right), showing how focus evolves across different model configurations. Each section includes original frames (first row), baseline model (second row), model with ShA adapters (third row), and full PE-CLIP with ShA and TDA (fourth row). Warmer colors indicate stronger attention, with PE-CLIP progressively refining focus on key facial regions, enhancing representation refinement (via ShA) and temporal modeling (via TDA).}
      \label{fig:attnmap}
  \end{figure*}

\noindent {\bf Attention Visualization:}
To "see" the contributions of PE-CLIP components, we visualize (in Figure~\ref{fig:attnmap}) the attention map of the last transformer block using the Gradient Attention Rollout technique, including sadness and anger. From top to bottom, the visualizations correspond to (1) the input (first row), (2) the baseline model (second row), (3) the baseline + ShA adapter (third row), and (4) the full PE-CLIP framework (Baseline + ShA + TDA) (fourth row). These visualizations show that PE-CLIP progressively refines attention allocation, enabling the model to focus on emotion-relevant facial regions while effectively capturing temporal variations across frames. For instance, 
In \textit{sadness}, we observe distinct improvements by adding PE-CLIP components. The baseline model displays widely dispersed attention, often focusing on irrelevant areas such as the cheeks and forehead. Adding ShA to both vision and text encoders, refines this distribution, directing the focus toward the brows, eyes, and mouth, which are strongly associated with AU1 (Inner Brow Raiser) and AU15 (Lip Corner Depressor), important indicators of sadness. With TDA integration, the attention becomes further concentrated, reinforcing the importance of these regions over time and capturing subtle temporal variations essential for distinguishing sadness, particularly the progressive furrowing of the brows across frames. A similar trend is observed for \textit{anger}, where the baseline model fails to localize important facial cues. Incorporating ShA shifts the attention toward the brows and jawline, aligning with AU4 (Brow Lowerer) and AU17 (Chin Raiser), key action units for anger. The TDA adapter further enhances this by allowing the model to track the increasing intensity of furrowed eyebrows and tightened lips over time, ensuring a more robust temporal understanding. These observations demonstrate that PE-CLIP effectively captures the dynamic transitions of facial expressions, refining attention to important regions and enhancing temporal consistency for robust DFER.

\begin{figure*}[!t]
      \centering
      \includegraphics[width=1.0\linewidth]{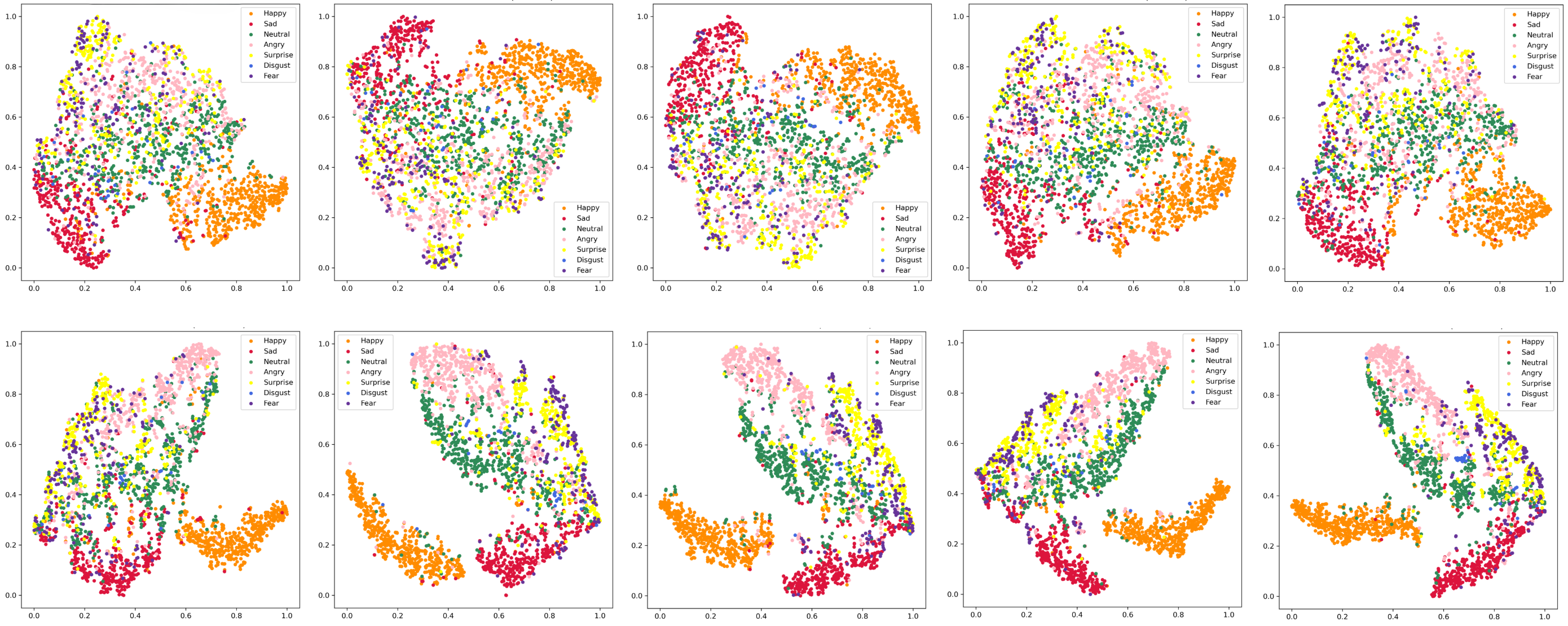}
      \caption{t-SNE Visualization of High-Level Feature Distributions on DFEW (fd1–5):  baseline Model, PE-CLIP without adapters and prompt  (Top row) vs. Proposed PE-CLIP model (Bottom row).}
      \label{fig:tsne}
  \end{figure*}

\vspace{3mm}  
\noindent {\bf Feature Distribution Visualization:}
Figure~\ref{fig:tsne} presents t-SNE (t-distributed stochastic neighbor embedding ~\cite{van2008visualizing}) visualizations comparing the feature distributions of the baseline model (top row) and PE-CLIP (bottom row) across five folds of the DFEW dataset. In the baseline model, features for visually similar emotions, such as \textit{fear} and \textit{surprise} (both involving raised eyebrows) or \textit{disgust} and \textit{anger} (where both characterized by furrowed brows and tightened lips), exhibit significant overlap, leading to higher misclassification rates.  Additionally, the baseline shows scattered and loosely clustered features, indicating limited temporal modeling and weak alignment between visual and textual modalities. In contrast, PE-CLIP demonstrates more well-separated and distinct clusters, reflecting its enhanced ability to capture discriminative features. Emotions with strong visual cues (\textit{happiness}, \textit{anger}, \textit{surprise}) exhibit well-defined clusters, while even challenging expressions like \textit{fear} and \textit{disgust} achieve improved separation. These visualizations further support the quantitative results of PE-CLIP and highlight the effectiveness of its temporal modeling (TDA) and feature refinement (ShA \& TA adapters) in generating semantically rich and discriminative representations. 
  

\section{Conclusion}
\label{sec:clc}
In this paper, we proposed PE-CLIP, a parameter-efficient method that effectively extended CLIP for DFER by integrating lightweight adapters and multi-modal prompt learning to improve temporal modeling and feature refinement, while requiring only 9M trainable parameters without full fine-tuning of CLIP’s encoders. The Temporal Dynamic Adapter (TDA), incorporating a GRU-based structure with dynamic scaling, allowed the model to adaptively emphasize significant temporal variations, while the Shared Adapter (ShA) 
facilitated efficient feature processing across both modalities. Additionaly, we leveraged multi-modal prompts (MaPLe) and Action Unit (AU)-based descriptions, providing semantically rich textual guidance to further improve vision-language consistency in dynamic facial expression recognition. Experimental results on benchmark datasets demonstrated that PE-CLIP achieved competitive performance compared to state-of-the-art methods while maintaining a lightweight design, making it a parameter-efficient and practical solution for DFER. Ablation study shed light into the contributions of each component of our proposed model, particularly in enhancing temporal modeling, features refinement, and optimizing the representation learning. This work can be further extended e.g. by exploring self-supervised learning techniques to enhance the generalization capabilities and reduce reliance on labeled datasets. It is also of interest to consider a multi-modal deep fusion framework, integrating audio-based emotion cues (in addition to text and visual cues) to further improve the robustness of dynamic facial expression recognition.


\bibliographystyle{ACM-Reference-Format}
\bibliography{sample-base}


\begin{thebibliography}{71}


\ifx \showCODEN    \undefined \def \showCODEN     #1{\unskip}     \fi
\ifx \showISBNx    \undefined \def \showISBNx     #1{\unskip}     \fi
\ifx \showISBNxiii \undefined \def \showISBNxiii  #1{\unskip}     \fi
\ifx \showISSN     \undefined \def \showISSN      #1{\unskip}     \fi
\ifx \showLCCN     \undefined \def \showLCCN      #1{\unskip}     \fi
\ifx \shownote     \undefined \def \shownote      #1{#1}          \fi
\ifx \showarticletitle \undefined \def \showarticletitle #1{#1}   \fi
\ifx \showURL      \undefined \def \showURL       {\relax}        \fi
\providecommand\bibfield[2]{#2}
\providecommand\bibinfo[2]{#2}
\providecommand\natexlab[1]{#1}
\providecommand\showeprint[2][]{arXiv:#2}

\bibitem[Ahonen et~al\mbox{.}(2006)]%
        {LBP}
\bibfield{author}{\bibinfo{person}{T. Ahonen}, \bibinfo{person}{A. Hadid}, {and} \bibinfo{person}{M. Pietikainen}.} \bibinfo{year}{2006}\natexlab{}.
\newblock \showarticletitle{Face Description with Local Binary Patterns: Application to Face Recognition}.
\newblock \bibinfo{journal}{\emph{IEEE Transactions on Pattern Analysis and Machine Intelligence}} \bibinfo{volume}{28}, \bibinfo{number}{12} (\bibinfo{year}{2006}), \bibinfo{pages}{2037--2041}.
\newblock
\href{https://doi.org/10.1109/TPAMI.2006.244}{doi:\nolinkurl{10.1109/TPAMI.2006.244}}


\bibitem[Alexey(2020)]%
        {alexey2020image}
\bibfield{author}{\bibinfo{person}{Dosovitskiy Alexey}.} \bibinfo{year}{2020}\natexlab{}.
\newblock \showarticletitle{An image is worth 16x16 words: Transformers for image recognition at scale}.
\newblock \bibinfo{journal}{\emph{arXiv preprint arXiv: 2010.11929}} (\bibinfo{year}{2020}).
\newblock


\bibitem[Cai et~al\mbox{.}(2023)]%
        {cai2023marlin}
\bibfield{author}{\bibinfo{person}{Zhixi Cai}, \bibinfo{person}{Shreya Ghosh}, \bibinfo{person}{Kalin Stefanov}, \bibinfo{person}{Abhinav Dhall}, \bibinfo{person}{Jianfei Cai}, \bibinfo{person}{Hamid Rezatofighi}, \bibinfo{person}{Reza Haffari}, {and} \bibinfo{person}{Munawar Hayat}.} \bibinfo{year}{2023}\natexlab{}.
\newblock \showarticletitle{Marlin: Masked autoencoder for facial video representation learning}. In \bibinfo{booktitle}{\emph{Proceedings of the IEEE/CVF conference on computer vision and pattern recognition}}. \bibinfo{pages}{1493--1504}.
\newblock


\bibitem[Chattopadhyay et~al\mbox{.}(2020)]%
        {chattopadhyay2020facial}
\bibfield{author}{\bibinfo{person}{Joyati Chattopadhyay}, \bibinfo{person}{Souvik Kundu}, \bibinfo{person}{Arpita Chakraborty}, {and} \bibinfo{person}{Jyoti~Sekhar Banerjee}.} \bibinfo{year}{2020}\natexlab{}.
\newblock \showarticletitle{Facial expression recognition for human computer interaction}.
\newblock \bibinfo{journal}{\emph{New Trends in Computational Vision and Bio-inspired Computing: Selected works presented at the ICCVBIC 2018, Coimbatore, India}} (\bibinfo{year}{2020}), \bibinfo{pages}{1181--1192}.
\newblock


\bibitem[Chen et~al\mbox{.}(2024a)]%
        {chen2024finecliper}
\bibfield{author}{\bibinfo{person}{Haodong Chen}, \bibinfo{person}{Haojian Huang}, \bibinfo{person}{Junhao Dong}, \bibinfo{person}{Mingzhe Zheng}, {and} \bibinfo{person}{Dian Shao}.} \bibinfo{year}{2024}\natexlab{a}.
\newblock \showarticletitle{Finecliper: Multi-modal fine-grained clip for dynamic facial expression recognition with adapters}. In \bibinfo{booktitle}{\emph{Proceedings of the 32nd ACM International Conference on Multimedia}}. \bibinfo{pages}{2301--2310}.
\newblock


\bibitem[Chen et~al\mbox{.}(2022)]%
        {chen2022adaptformer}
\bibfield{author}{\bibinfo{person}{Shoufa Chen}, \bibinfo{person}{Chongjian Ge}, \bibinfo{person}{Zhan Tong}, \bibinfo{person}{Jiangliu Wang}, \bibinfo{person}{Yibing Song}, \bibinfo{person}{Jue Wang}, {and} \bibinfo{person}{Ping Luo}.} \bibinfo{year}{2022}\natexlab{}.
\newblock \showarticletitle{Adaptformer: Adapting vision transformers for scalable visual recognition}.
\newblock \bibinfo{journal}{\emph{Advances in Neural Information Processing Systems}}  \bibinfo{volume}{35} (\bibinfo{year}{2022}), \bibinfo{pages}{16664--16678}.
\newblock


\bibitem[Chen et~al\mbox{.}(2024b)]%
        {chen2024static}
\bibfield{author}{\bibinfo{person}{Yin Chen}, \bibinfo{person}{Jia Li}, \bibinfo{person}{Shiguang Shan}, \bibinfo{person}{Meng Wang}, {and} \bibinfo{person}{Richang Hong}.} \bibinfo{year}{2024}\natexlab{b}.
\newblock \showarticletitle{From static to dynamic: Adapting landmark-aware image models for facial expression recognition in videos}.
\newblock \bibinfo{journal}{\emph{IEEE Transactions on Affective Computing}} (\bibinfo{year}{2024}).
\newblock


\bibitem[Cho et~al\mbox{.}(2014)]%
        {cho2014learning}
\bibfield{author}{\bibinfo{person}{Kyunghyun Cho}, \bibinfo{person}{Bart Van~Merri{\"e}nboer}, \bibinfo{person}{Caglar Gulcehre}, \bibinfo{person}{Dzmitry Bahdanau}, \bibinfo{person}{Fethi Bougares}, \bibinfo{person}{Holger Schwenk}, {and} \bibinfo{person}{Yoshua Bengio}.} \bibinfo{year}{2014}\natexlab{}.
\newblock \showarticletitle{Learning phrase representations using RNN encoder-decoder for statistical machine translation}.
\newblock \bibinfo{journal}{\emph{arXiv preprint arXiv:1406.1078}} (\bibinfo{year}{2014}).
\newblock


\bibitem[Chung et~al\mbox{.}(2014)]%
        {chung2014empirical}
\bibfield{author}{\bibinfo{person}{Junyoung Chung}, \bibinfo{person}{Caglar Gulcehre}, \bibinfo{person}{KyungHyun Cho}, {and} \bibinfo{person}{Yoshua Bengio}.} \bibinfo{year}{2014}\natexlab{}.
\newblock \showarticletitle{Empirical evaluation of gated recurrent neural networks on sequence modeling}.
\newblock \bibinfo{journal}{\emph{arXiv preprint arXiv:1412.3555}} (\bibinfo{year}{2014}).
\newblock


\bibitem[Dosovitskiy et~al\mbox{.}(2020)]%
        {dosovitskiy2020image}
\bibfield{author}{\bibinfo{person}{Alexey Dosovitskiy}, \bibinfo{person}{Lucas Beyer}, \bibinfo{person}{Alexander Kolesnikov}, \bibinfo{person}{Dirk Weissenborn}, \bibinfo{person}{Xiaohua Zhai}, \bibinfo{person}{Thomas Unterthiner}, \bibinfo{person}{Mostafa Dehghani}, \bibinfo{person}{Matthias Minderer}, \bibinfo{person}{Georg Heigold}, \bibinfo{person}{Sylvain Gelly}, {et~al\mbox{.}}} \bibinfo{year}{2020}\natexlab{}.
\newblock \showarticletitle{An image is worth 16x16 words: Transformers for image recognition at scale}.
\newblock \bibinfo{journal}{\emph{arXiv preprint arXiv:2010.11929}} (\bibinfo{year}{2020}).
\newblock


\bibitem[Du et~al\mbox{.}(2014)]%
        {du2014compound}
\bibfield{author}{\bibinfo{person}{Shichuan Du}, \bibinfo{person}{Yong Tao}, {and} \bibinfo{person}{Aleix~M Martinez}.} \bibinfo{year}{2014}\natexlab{}.
\newblock \showarticletitle{Compound facial expressions of emotion}.
\newblock \bibinfo{journal}{\emph{Proceedings of the national academy of sciences}} \bibinfo{volume}{111}, \bibinfo{number}{15} (\bibinfo{year}{2014}), \bibinfo{pages}{E1454--E1462}.
\newblock


\bibitem[Ekman and Rosenberg(1997)]%
        {ekman1997face}
\bibfield{author}{\bibinfo{person}{Paul Ekman} {and} \bibinfo{person}{Erika~L Rosenberg}.} \bibinfo{year}{1997}\natexlab{}.
\newblock \bibinfo{booktitle}{\emph{What the face reveals: Basic and applied studies of spontaneous expression using the Facial Action Coding System (FACS)}}.
\newblock \bibinfo{publisher}{Oxford University Press, USA}.
\newblock


\bibitem[Fan et~al\mbox{.}(2016)]%
        {10.1145/2993148.2997632}
\bibfield{author}{\bibinfo{person}{Yin Fan}, \bibinfo{person}{Xiangju Lu}, \bibinfo{person}{Dian Li}, {and} \bibinfo{person}{Yuanliu Liu}.} \bibinfo{year}{2016}\natexlab{}.
\newblock \showarticletitle{Video-based emotion recognition using CNN-RNN and C3D hybrid networks}. In \bibinfo{booktitle}{\emph{Proceedings of the 18th ACM International Conference on Multimodal Interaction}} (Tokyo, Japan) \emph{(\bibinfo{series}{ICMI '16})}. \bibinfo{publisher}{Association for Computing Machinery}, \bibinfo{address}{New York, NY, USA}, \bibinfo{pages}{445–450}.
\newblock
\showISBNx{9781450345569}
\href{https://doi.org/10.1145/2993148.2997632}{doi:\nolinkurl{10.1145/2993148.2997632}}


\bibitem[Foteinopoulou and Patras(2024)]%
        {foteinopoulou2024emoclip}
\bibfield{author}{\bibinfo{person}{Niki~Maria Foteinopoulou} {and} \bibinfo{person}{Ioannis Patras}.} \bibinfo{year}{2024}\natexlab{}.
\newblock \showarticletitle{Emoclip: A vision-language method for zero-shot video facial expression recognition}. In \bibinfo{booktitle}{\emph{2024 IEEE 18th International Conference on Automatic Face and Gesture Recognition (FG)}}. IEEE, \bibinfo{pages}{1--10}.
\newblock


\bibitem[Ghosh et~al\mbox{.}(2024)]%
        {ghosh2024clipsyntel}
\bibfield{author}{\bibinfo{person}{Akash Ghosh}, \bibinfo{person}{Arkadeep Acharya}, \bibinfo{person}{Raghav Jain}, \bibinfo{person}{Sriparna Saha}, \bibinfo{person}{Aman Chadha}, {and} \bibinfo{person}{Setu Sinha}.} \bibinfo{year}{2024}\natexlab{}.
\newblock \showarticletitle{Clipsyntel: clip and llm synergy for multimodal question summarization in healthcare}. In \bibinfo{booktitle}{\emph{Proceedings of the AAAI Conference on Artificial Intelligence}}, Vol.~\bibinfo{volume}{38}. \bibinfo{pages}{22031--22039}.
\newblock


\bibitem[Gowda et~al\mbox{.}(2024)]%
        {gowda2024fe}
\bibfield{author}{\bibinfo{person}{Shreyank~N Gowda}, \bibinfo{person}{Boyan Gao}, {and} \bibinfo{person}{David~A Clifton}.} \bibinfo{year}{2024}\natexlab{}.
\newblock \showarticletitle{Fe-adapter: Adapting image-based emotion classifiers to videos}. In \bibinfo{booktitle}{\emph{2024 IEEE 18th International Conference on Automatic Face and Gesture Recognition (FG)}}. IEEE, \bibinfo{pages}{1--6}.
\newblock


\bibitem[Han and Li(2024)]%
        {10743836}
\bibfield{author}{\bibinfo{person}{Yitao Han} {and} \bibinfo{person}{Qiming Li}.} \bibinfo{year}{2024}\natexlab{}.
\newblock \showarticletitle{DFLM: A Dynamic Facial-Language Model Based on CLIP}. In \bibinfo{booktitle}{\emph{2024 9th International Conference on Intelligent Computing and Signal Processing (ICSP)}}. \bibinfo{pages}{1132--1137}.
\newblock
\href{https://doi.org/10.1109/ICSP62122.2024.10743836}{doi:\nolinkurl{10.1109/ICSP62122.2024.10743836}}


\bibitem[Harris et~al\mbox{.}(2020)]%
        {harris2020fmix}
\bibfield{author}{\bibinfo{person}{Ethan Harris}, \bibinfo{person}{Antonia Marcu}, \bibinfo{person}{Matthew Painter}, \bibinfo{person}{Mahesan Niranjan}, \bibinfo{person}{Adam Pr{\"u}gel-Bennett}, {and} \bibinfo{person}{Jonathon Hare}.} \bibinfo{year}{2020}\natexlab{}.
\newblock \showarticletitle{Fmix: Enhancing mixed sample data augmentation}.
\newblock \bibinfo{journal}{\emph{arXiv preprint arXiv:2002.12047}} (\bibinfo{year}{2020}).
\newblock


\bibitem[He et~al\mbox{.}(2021)]%
        {he2021towards}
\bibfield{author}{\bibinfo{person}{Junxian He}, \bibinfo{person}{Chunting Zhou}, \bibinfo{person}{Xuezhe Ma}, \bibinfo{person}{Taylor Berg-Kirkpatrick}, {and} \bibinfo{person}{Graham Neubig}.} \bibinfo{year}{2021}\natexlab{}.
\newblock \showarticletitle{Towards a unified view of parameter-efficient transfer learning}.
\newblock \bibinfo{journal}{\emph{arXiv preprint arXiv:2110.04366}} (\bibinfo{year}{2021}).
\newblock


\bibitem[He et~al\mbox{.}(2016)]%
        {he2016deep}
\bibfield{author}{\bibinfo{person}{Kaiming He}, \bibinfo{person}{Xiangyu Zhang}, \bibinfo{person}{Shaoqing Ren}, {and} \bibinfo{person}{Jian Sun}.} \bibinfo{year}{2016}\natexlab{}.
\newblock \showarticletitle{Deep residual learning for image recognition}. In \bibinfo{booktitle}{\emph{Proceedings of the IEEE conference on computer vision and pattern recognition}}. \bibinfo{pages}{770--778}.
\newblock


\bibitem[Hochreiter(1997)]%
        {hochreiter1997long}
\bibfield{author}{\bibinfo{person}{S Hochreiter}.} \bibinfo{year}{1997}\natexlab{}.
\newblock \showarticletitle{Long Short-term Memory}.
\newblock \bibinfo{journal}{\emph{Neural Computation MIT-Press}} (\bibinfo{year}{1997}).
\newblock


\bibitem[Houlsby et~al\mbox{.}(2019)]%
        {houlsby2019parameter}
\bibfield{author}{\bibinfo{person}{Neil Houlsby}, \bibinfo{person}{Andrei Giurgiu}, \bibinfo{person}{Stanislaw Jastrzebski}, \bibinfo{person}{Bruna Morrone}, \bibinfo{person}{Quentin De~Laroussilhe}, \bibinfo{person}{Andrea Gesmundo}, \bibinfo{person}{Mona Attariyan}, {and} \bibinfo{person}{Sylvain Gelly}.} \bibinfo{year}{2019}\natexlab{}.
\newblock \showarticletitle{Parameter-efficient transfer learning for NLP}. In \bibinfo{booktitle}{\emph{International conference on machine learning}}. PMLR, \bibinfo{pages}{2790--2799}.
\newblock


\bibitem[Jia et~al\mbox{.}(2022)]%
        {jia2022visual}
\bibfield{author}{\bibinfo{person}{Menglin Jia}, \bibinfo{person}{Luming Tang}, \bibinfo{person}{Bor-Chun Chen}, \bibinfo{person}{Claire Cardie}, \bibinfo{person}{Serge Belongie}, \bibinfo{person}{Bharath Hariharan}, {and} \bibinfo{person}{Ser-Nam Lim}.} \bibinfo{year}{2022}\natexlab{}.
\newblock \showarticletitle{Visual prompt tuning}. In \bibinfo{booktitle}{\emph{European Conference on Computer Vision}}. Springer, \bibinfo{pages}{709--727}.
\newblock


\bibitem[Jiang et~al\mbox{.}(2020)]%
        {jiang2020dfew}
\bibfield{author}{\bibinfo{person}{Xingxun Jiang}, \bibinfo{person}{Yuan Zong}, \bibinfo{person}{Wenming Zheng}, \bibinfo{person}{Chuangao Tang}, \bibinfo{person}{Wanchuang Xia}, \bibinfo{person}{Cheng Lu}, {and} \bibinfo{person}{Jiateng Liu}.} \bibinfo{year}{2020}\natexlab{}.
\newblock \showarticletitle{Dfew: A large-scale database for recognizing dynamic facial expressions in the wild}. In \bibinfo{booktitle}{\emph{Proceedings of the 28th ACM international conference on multimedia}}. \bibinfo{pages}{2881--2889}.
\newblock


\bibitem[Jin et~al\mbox{.}(2024)]%
        {jin2024transformer}
\bibfield{author}{\bibinfo{person}{Xing Jin}, \bibinfo{person}{Xulin Song}, \bibinfo{person}{Xiyin Wu}, {and} \bibinfo{person}{Wenzhu Yan}.} \bibinfo{year}{2024}\natexlab{}.
\newblock \showarticletitle{Transformer embedded spectral-based graph network for facial expression recognition}.
\newblock \bibinfo{journal}{\emph{International Journal of Machine Learning and Cybernetics}} \bibinfo{volume}{15}, \bibinfo{number}{6} (\bibinfo{year}{2024}), \bibinfo{pages}{2063--2077}.
\newblock


\bibitem[Khanna et~al\mbox{.}(2024)]%
        {khanna2024enhanced}
\bibfield{author}{\bibinfo{person}{Deepanshu Khanna}, \bibinfo{person}{Neeru Jindal}, \bibinfo{person}{Prashant~Singh Rana}, {and} \bibinfo{person}{Harpreet Singh}.} \bibinfo{year}{2024}\natexlab{}.
\newblock \showarticletitle{Enhanced spatio-temporal 3D CNN for facial expression classification in videos}.
\newblock \bibinfo{journal}{\emph{Multimedia Tools and Applications}} \bibinfo{volume}{83}, \bibinfo{number}{4} (\bibinfo{year}{2024}), \bibinfo{pages}{9911--9928}.
\newblock


\bibitem[Khattak et~al\mbox{.}(2023)]%
        {khattak2023maple}
\bibfield{author}{\bibinfo{person}{Muhammad~Uzair Khattak}, \bibinfo{person}{Hanoona Rasheed}, \bibinfo{person}{Muhammad Maaz}, \bibinfo{person}{Salman Khan}, {and} \bibinfo{person}{Fahad~Shahbaz Khan}.} \bibinfo{year}{2023}\natexlab{}.
\newblock \showarticletitle{Maple: Multi-modal prompt learning}. In \bibinfo{booktitle}{\emph{Proceedings of the IEEE/CVF Conference on Computer Vision and Pattern Recognition}}. \bibinfo{pages}{19113--19122}.
\newblock


\bibitem[Lee et~al\mbox{.}(2023)]%
        {lee2023frame}
\bibfield{author}{\bibinfo{person}{Bokyeung Lee}, \bibinfo{person}{Hyunuk Shin}, \bibinfo{person}{Bonhwa Ku}, {and} \bibinfo{person}{Hanseok Ko}.} \bibinfo{year}{2023}\natexlab{}.
\newblock \showarticletitle{Frame level emotion guided dynamic facial expression recognition with emotion grouping}. In \bibinfo{booktitle}{\emph{Proceedings of the IEEE/CVF Conference on Computer Vision and Pattern Recognition}}. \bibinfo{pages}{5681--5691}.
\newblock


\bibitem[Lee et~al\mbox{.}(2020)]%
        {9102419}
\bibfield{author}{\bibinfo{person}{Jiyoung Lee}, \bibinfo{person}{Sunok Kim}, \bibinfo{person}{Seungryong Kim}, {and} \bibinfo{person}{Kwanghoon Sohn}.} \bibinfo{year}{2020}\natexlab{}.
\newblock \showarticletitle{Multi-Modal Recurrent Attention Networks for Facial Expression Recognition}.
\newblock \bibinfo{journal}{\emph{IEEE Transactions on Image Processing}}  \bibinfo{volume}{29} (\bibinfo{year}{2020}), \bibinfo{pages}{6977--6991}.
\newblock
\href{https://doi.org/10.1109/TIP.2020.2996086}{doi:\nolinkurl{10.1109/TIP.2020.2996086}}


\bibitem[Li et~al\mbox{.}(2023)]%
        {li2023intensity}
\bibfield{author}{\bibinfo{person}{Hanting Li}, \bibinfo{person}{Hongjing Niu}, \bibinfo{person}{Zhaoqing Zhu}, {and} \bibinfo{person}{Feng Zhao}.} \bibinfo{year}{2023}\natexlab{}.
\newblock \showarticletitle{Intensity-aware loss for dynamic facial expression recognition in the wild}. In \bibinfo{booktitle}{\emph{Proceedings of the AAAI Conference on Artificial Intelligence}}, Vol.~\bibinfo{volume}{37}. \bibinfo{pages}{67--75}.
\newblock


\bibitem[Li et~al\mbox{.}(2024a)]%
        {li2024cliper}
\bibfield{author}{\bibinfo{person}{Hanting Li}, \bibinfo{person}{Hongjing Niu}, \bibinfo{person}{Zhaoqing Zhu}, {and} \bibinfo{person}{Feng Zhao}.} \bibinfo{year}{2024}\natexlab{a}.
\newblock \showarticletitle{Cliper: A unified vision-language framework for in-the-wild facial expression recognition}. In \bibinfo{booktitle}{\emph{2024 IEEE International Conference on Multimedia and Expo (ICME)}}. IEEE, \bibinfo{pages}{1--6}.
\newblock


\bibitem[Li et~al\mbox{.}(2022)]%
        {li2022nr}
\bibfield{author}{\bibinfo{person}{Hanting Li}, \bibinfo{person}{Mingzhe Sui}, \bibinfo{person}{Zhaoqing Zhu}, {et~al\mbox{.}}} \bibinfo{year}{2022}\natexlab{}.
\newblock \showarticletitle{Nr-dfernet: Noise-robust network for dynamic facial expression recognition}.
\newblock \bibinfo{journal}{\emph{arXiv preprint arXiv:2206.04975}} (\bibinfo{year}{2022}).
\newblock


\bibitem[Li et~al\mbox{.}(2024b)]%
        {li2024dual}
\bibfield{author}{\bibinfo{person}{Min Li}, \bibinfo{person}{Xiaoqin Zhang}, \bibinfo{person}{Chenxiang Fan}, \bibinfo{person}{Tangfei Liao}, {and} \bibinfo{person}{Guobao Xiao}.} \bibinfo{year}{2024}\natexlab{b}.
\newblock \showarticletitle{Dual-STI: Dual-path Spatial-Temporal Interaction Learning for Dynamic Facial Expression Recognition}.
\newblock \bibinfo{journal}{\emph{Information Sciences}} (\bibinfo{year}{2024}), \bibinfo{pages}{120953}.
\newblock


\bibitem[Lin et~al\mbox{.}(2024)]%
        {lin2024rethinking}
\bibfield{author}{\bibinfo{person}{Kun-Yu Lin}, \bibinfo{person}{Henghui Ding}, \bibinfo{person}{Jiaming Zhou}, \bibinfo{person}{Yu-Ming Tang}, \bibinfo{person}{Yi-Xing Peng}, \bibinfo{person}{Zhilin Zhao}, \bibinfo{person}{Chen~Change Loy}, {and} \bibinfo{person}{Wei-Shi Zheng}.} \bibinfo{year}{2024}\natexlab{}.
\newblock \showarticletitle{Rethinking clip-based video learners in cross-domain open-vocabulary action recognition}.
\newblock \bibinfo{journal}{\emph{arXiv preprint arXiv:2403.01560}} (\bibinfo{year}{2024}).
\newblock


\bibitem[Liu et~al\mbox{.}(2022a)]%
        {liu2022mafw}
\bibfield{author}{\bibinfo{person}{Yuanyuan Liu}, \bibinfo{person}{Wei Dai}, \bibinfo{person}{Chuanxu Feng}, \bibinfo{person}{Wenbin Wang}, \bibinfo{person}{Guanghao Yin}, \bibinfo{person}{Jiabei Zeng}, {and} \bibinfo{person}{Shiguang Shan}.} \bibinfo{year}{2022}\natexlab{a}.
\newblock \showarticletitle{Mafw: A large-scale, multi-modal, compound affective database for dynamic facial expression recognition in the wild}. In \bibinfo{booktitle}{\emph{Proceedings of the 30th ACM International Conference on Multimedia}}. \bibinfo{pages}{24--32}.
\newblock


\bibitem[Liu et~al\mbox{.}(2022b)]%
        {liu2022clip}
\bibfield{author}{\bibinfo{person}{Yuanyuan Liu}, \bibinfo{person}{Chuanxu Feng}, \bibinfo{person}{Xiaohui Yuan}, \bibinfo{person}{Lin Zhou}, \bibinfo{person}{Wenbin Wang}, \bibinfo{person}{Jie Qin}, {and} \bibinfo{person}{Zhongwen Luo}.} \bibinfo{year}{2022}\natexlab{b}.
\newblock \showarticletitle{Clip-aware expressive feature learning for video-based facial expression recognition}.
\newblock \bibinfo{journal}{\emph{Information Sciences}}  \bibinfo{volume}{598} (\bibinfo{year}{2022}), \bibinfo{pages}{182--195}.
\newblock


\bibitem[Liu et~al\mbox{.}(2023)]%
        {liu2023expression}
\bibfield{author}{\bibinfo{person}{Yuanyuan Liu}, \bibinfo{person}{Wenbin Wang}, \bibinfo{person}{Chuanxu Feng}, \bibinfo{person}{Haoyu Zhang}, \bibinfo{person}{Zhe Chen}, {and} \bibinfo{person}{Yibing Zhan}.} \bibinfo{year}{2023}\natexlab{}.
\newblock \showarticletitle{Expression snippet transformer for robust video-based facial expression recognition}.
\newblock \bibinfo{journal}{\emph{Pattern Recognition}}  \bibinfo{volume}{138} (\bibinfo{year}{2023}), \bibinfo{pages}{109368}.
\newblock


\bibitem[Livingstone and Russo(2018)]%
        {livingstone2018ryerson}
\bibfield{author}{\bibinfo{person}{Steven~R Livingstone} {and} \bibinfo{person}{Frank~A Russo}.} \bibinfo{year}{2018}\natexlab{}.
\newblock \showarticletitle{The Ryerson Audio-Visual Database of Emotional Speech and Song (RAVDESS): A dynamic, multimodal set of facial and vocal expressions in North American English}.
\newblock \bibinfo{journal}{\emph{PloS one}} \bibinfo{volume}{13}, \bibinfo{number}{5} (\bibinfo{year}{2018}), \bibinfo{pages}{e0196391}.
\newblock


\bibitem[Ma et~al\mbox{.}(2022)]%
        {ma2022spatio}
\bibfield{author}{\bibinfo{person}{Fuyan Ma}, \bibinfo{person}{Bin Sun}, {and} \bibinfo{person}{Shutao Li}.} \bibinfo{year}{2022}\natexlab{}.
\newblock \showarticletitle{Spatio-temporal transformer for dynamic facial expression recognition in the wild}.
\newblock \bibinfo{journal}{\emph{arXiv preprint arXiv:2205.04749}} (\bibinfo{year}{2022}).
\newblock


\bibitem[Ma et~al\mbox{.}(2023)]%
        {ma2023logo}
\bibfield{author}{\bibinfo{person}{Fuyan Ma}, \bibinfo{person}{Bin Sun}, {and} \bibinfo{person}{Shutao Li}.} \bibinfo{year}{2023}\natexlab{}.
\newblock \showarticletitle{Logo-former: Local-global spatio-temporal transformer for dynamic facial expression recognition}. In \bibinfo{booktitle}{\emph{ICASSP 2023-2023 IEEE International Conference on Acoustics, Speech and Signal Processing (ICASSP)}}. IEEE, \bibinfo{pages}{1--5}.
\newblock


\bibitem[Manalu and Rifai(2024)]%
        {manalu2024detection}
\bibfield{author}{\bibinfo{person}{Haposan~Vincentius Manalu} {and} \bibinfo{person}{Achmad~Pratama Rifai}.} \bibinfo{year}{2024}\natexlab{}.
\newblock \showarticletitle{Detection of human emotions through facial expressions using hybrid convolutional neural network-recurrent neural network algorithm}.
\newblock \bibinfo{journal}{\emph{Intelligent Systems with Applications}}  \bibinfo{volume}{21} (\bibinfo{year}{2024}), \bibinfo{pages}{200339}.
\newblock


\bibitem[Radford et~al\mbox{.}(2021)]%
        {radford2021learning}
\bibfield{author}{\bibinfo{person}{Alec Radford}, \bibinfo{person}{Jong~Wook Kim}, \bibinfo{person}{Chris Hallacy}, \bibinfo{person}{Aditya Ramesh}, \bibinfo{person}{Gabriel Goh}, \bibinfo{person}{Sandhini Agarwal}, \bibinfo{person}{Girish Sastry}, \bibinfo{person}{Amanda Askell}, \bibinfo{person}{Pamela Mishkin}, \bibinfo{person}{Jack Clark}, {et~al\mbox{.}}} \bibinfo{year}{2021}\natexlab{}.
\newblock \showarticletitle{Learning transferable visual models from natural language supervision}. In \bibinfo{booktitle}{\emph{International conference on machine learning}}. PMLR, \bibinfo{pages}{8748--8763}.
\newblock


\bibitem[Ren et~al\mbox{.}(2024)]%
        {ren2024facial}
\bibfield{author}{\bibinfo{person}{Weihong Ren}, \bibinfo{person}{Yu Gao}, \bibinfo{person}{Zhi Han}, \bibinfo{person}{Zhiyong Wang}, \bibinfo{person}{Jiaole Wang}, \bibinfo{person}{Honghai Liu}, {et~al\mbox{.}}} \bibinfo{year}{2024}\natexlab{}.
\newblock \showarticletitle{Facial Expression Monitoring via Fine-Grained Vision-Language Alignment}.
\newblock \bibinfo{journal}{\emph{IEEE Transactions on Automation Science and Engineering}} (\bibinfo{year}{2024}).
\newblock


\bibitem[Saadi et~al\mbox{.}(2023)]%
        {saadi2023driver}
\bibfield{author}{\bibinfo{person}{Ibtissam Saadi}, \bibinfo{person}{Douglas~W Cunningham}, \bibinfo{person}{Taleb-Ahmed Abdelmalik}, \bibinfo{person}{Abdenour Hadid}, {and} \bibinfo{person}{Yassin El~Hillali}.} \bibinfo{year}{2023}\natexlab{}.
\newblock \showarticletitle{Driver’s facial expression recognition: A comprehensive survey}.
\newblock \bibinfo{journal}{\emph{Expert Systems with Applications}} (\bibinfo{year}{2023}), \bibinfo{pages}{122784}.
\newblock


\bibitem[Saadi et~al\mbox{.}(2025)]%
        {10.1007/978-3-031-80856-2_6}
\bibfield{author}{\bibinfo{person}{Ibtissam Saadi}, \bibinfo{person}{Abdenour Hadid}, \bibinfo{person}{Douglas~W. Cunningham}, \bibinfo{person}{Abdelmalik Taleb-Ahmed}, {and} \bibinfo{person}{Yassin El~Hillali}.} \bibinfo{year}{2025}\natexlab{}.
\newblock \showarticletitle{Leveraging Vision Language Models for Facial Expression Recognition in Driving Environment}. In \bibinfo{booktitle}{\emph{Sensor-Based Activity Recognition and Artificial Intelligence}}, \bibfield{editor}{\bibinfo{person}{Orhan Konak}, \bibinfo{person}{Bert Arnrich}, \bibinfo{person}{Gerald Bieber}, \bibinfo{person}{Arjan Kuijper}, {and} \bibinfo{person}{Sebastian Fudickar}} (Eds.). \bibinfo{publisher}{Springer Nature Switzerland}, \bibinfo{address}{Cham}, \bibinfo{pages}{81--93}.
\newblock
\showISBNx{978-3-031-80856-2}


\bibitem[Simonyan(2014)]%
        {simonyan2014very}
\bibfield{author}{\bibinfo{person}{Karen Simonyan}.} \bibinfo{year}{2014}\natexlab{}.
\newblock \showarticletitle{Very deep convolutional networks for large-scale image recognition}.
\newblock \bibinfo{journal}{\emph{arXiv preprint arXiv:1409.1556}} (\bibinfo{year}{2014}).
\newblock


\bibitem[Sun et~al\mbox{.}(2023)]%
        {sun2023mae}
\bibfield{author}{\bibinfo{person}{Licai Sun}, \bibinfo{person}{Zheng Lian}, \bibinfo{person}{Bin Liu}, {and} \bibinfo{person}{Jianhua Tao}.} \bibinfo{year}{2023}\natexlab{}.
\newblock \showarticletitle{Mae-dfer: Efficient masked autoencoder for self-supervised dynamic facial expression recognition}. In \bibinfo{booktitle}{\emph{Proceedings of the 31st ACM International Conference on Multimedia}}. \bibinfo{pages}{6110--6121}.
\newblock


\bibitem[Sun et~al\mbox{.}(2020)]%
        {sun2020multi}
\bibfield{author}{\bibinfo{person}{Licai Sun}, \bibinfo{person}{Zheng Lian}, \bibinfo{person}{Jianhua Tao}, \bibinfo{person}{Bin Liu}, {and} \bibinfo{person}{Mingyue Niu}.} \bibinfo{year}{2020}\natexlab{}.
\newblock \showarticletitle{Multi-modal continuous dimensional emotion recognition using recurrent neural network and self-attention mechanism}. In \bibinfo{booktitle}{\emph{Proceedings of the 1st international on multimodal sentiment analysis in real-life media challenge and workshop}}. \bibinfo{pages}{27--34}.
\newblock


\bibitem[Tao et~al\mbox{.}(2023)]%
        {tao2023freq}
\bibfield{author}{\bibinfo{person}{Zeng Tao}, \bibinfo{person}{Yan Wang}, \bibinfo{person}{Zhaoyu Chen}, \bibinfo{person}{Boyang Wang}, \bibinfo{person}{Shaoqi Yan}, \bibinfo{person}{Kaixun Jiang}, \bibinfo{person}{Shuyong Gao}, {and} \bibinfo{person}{Wenqiang Zhang}.} \bibinfo{year}{2023}\natexlab{}.
\newblock \showarticletitle{Freq-hd: An interpretable frequency-based high-dynamics affective clip selection method for in-the-wild facial expression recognition in videos}. In \bibinfo{booktitle}{\emph{Proceedings of the 31st ACM International Conference on Multimedia}}. \bibinfo{pages}{843--852}.
\newblock


\bibitem[Tran et~al\mbox{.}(2015)]%
        {7410867}
\bibfield{author}{\bibinfo{person}{Du Tran}, \bibinfo{person}{Lubomir Bourdev}, \bibinfo{person}{Rob Fergus}, \bibinfo{person}{Lorenzo Torresani}, {and} \bibinfo{person}{Manohar Paluri}.} \bibinfo{year}{2015}\natexlab{}.
\newblock \showarticletitle{Learning Spatiotemporal Features with 3D Convolutional Networks}. In \bibinfo{booktitle}{\emph{2015 IEEE International Conference on Computer Vision (ICCV)}}. \bibinfo{pages}{4489--4497}.
\newblock
\href{https://doi.org/10.1109/ICCV.2015.510}{doi:\nolinkurl{10.1109/ICCV.2015.510}}


\bibitem[Van~der Maaten and Hinton(2008)]%
        {van2008visualizing}
\bibfield{author}{\bibinfo{person}{Laurens Van~der Maaten} {and} \bibinfo{person}{Geoffrey Hinton}.} \bibinfo{year}{2008}\natexlab{}.
\newblock \showarticletitle{Visualizing data using t-SNE.}
\newblock \bibinfo{journal}{\emph{Journal of machine learning research}} \bibinfo{volume}{9}, \bibinfo{number}{11} (\bibinfo{year}{2008}).
\newblock


\bibitem[Wang et~al\mbox{.}(2023a)]%
        {10204167}
\bibfield{author}{\bibinfo{person}{Hanyang Wang}, \bibinfo{person}{Bo Li}, \bibinfo{person}{Shuang Wu}, \bibinfo{person}{Siyuan Shen}, \bibinfo{person}{Feng Liu}, \bibinfo{person}{Shouhong Ding}, {and} \bibinfo{person}{Aimin Zhou}.} \bibinfo{year}{2023}\natexlab{a}.
\newblock \showarticletitle{Rethinking the Learning Paradigm for Dynamic Facial Expression Recognition}. In \bibinfo{booktitle}{\emph{2023 IEEE/CVF Conference on Computer Vision and Pattern Recognition (CVPR)}}. \bibinfo{pages}{17958--17968}.
\newblock
\href{https://doi.org/10.1109/CVPR52729.2023.01722}{doi:\nolinkurl{10.1109/CVPR52729.2023.01722}}


\bibitem[Wang et~al\mbox{.}(2023b)]%
        {wang2023rethinking}
\bibfield{author}{\bibinfo{person}{Hanyang Wang}, \bibinfo{person}{Bo Li}, \bibinfo{person}{Shuang Wu}, \bibinfo{person}{Siyuan Shen}, \bibinfo{person}{Feng Liu}, \bibinfo{person}{Shouhong Ding}, {and} \bibinfo{person}{Aimin Zhou}.} \bibinfo{year}{2023}\natexlab{b}.
\newblock \showarticletitle{Rethinking the learning paradigm for dynamic facial expression recognition}. In \bibinfo{booktitle}{\emph{Proceedings of the IEEE/CVF conference on computer vision and pattern recognition}}. \bibinfo{pages}{17958--17968}.
\newblock


\bibitem[Wang et~al\mbox{.}(2022a)]%
        {wang2022ferv39k}
\bibfield{author}{\bibinfo{person}{Yan Wang}, \bibinfo{person}{Yixuan Sun}, \bibinfo{person}{Yiwen Huang}, \bibinfo{person}{Zhongying Liu}, \bibinfo{person}{Shuyong Gao}, \bibinfo{person}{Wei Zhang}, \bibinfo{person}{Weifeng Ge}, {and} \bibinfo{person}{Wenqiang Zhang}.} \bibinfo{year}{2022}\natexlab{a}.
\newblock \showarticletitle{Ferv39k: A large-scale multi-scene dataset for facial expression recognition in videos}. In \bibinfo{booktitle}{\emph{Proceedings of the IEEE/CVF conference on computer vision and pattern recognition}}. \bibinfo{pages}{20922--20931}.
\newblock


\bibitem[Wang et~al\mbox{.}(2022b)]%
        {wang2022dpcnet}
\bibfield{author}{\bibinfo{person}{Yan Wang}, \bibinfo{person}{Yixuan Sun}, \bibinfo{person}{Wei Song}, \bibinfo{person}{Shuyong Gao}, \bibinfo{person}{Yiwen Huang}, \bibinfo{person}{Zhaoyu Chen}, \bibinfo{person}{Weifeng Ge}, {and} \bibinfo{person}{Wenqiang Zhang}.} \bibinfo{year}{2022}\natexlab{b}.
\newblock \showarticletitle{Dpcnet: Dual path multi-excitation collaborative network for facial expression representation learning in videos}. In \bibinfo{booktitle}{\emph{Proceedings of the 30th ACM International Conference on Multimedia}}. \bibinfo{pages}{101--110}.
\newblock


\bibitem[Xia and Jiang(2023)]%
        {xia2023hit}
\bibfield{author}{\bibinfo{person}{Xiaohan Xia} {and} \bibinfo{person}{Dongmei Jiang}.} \bibinfo{year}{2023}\natexlab{}.
\newblock \showarticletitle{HiT-MST: Dynamic facial expression recognition with hierarchical transformers and multi-scale spatiotemporal aggregation}.
\newblock \bibinfo{journal}{\emph{Information Sciences}}  \bibinfo{volume}{644} (\bibinfo{year}{2023}), \bibinfo{pages}{119301}.
\newblock


\bibitem[Xin et~al\mbox{.}(2024)]%
        {xin2024vmt}
\bibfield{author}{\bibinfo{person}{Yi Xin}, \bibinfo{person}{Junlong Du}, \bibinfo{person}{Qiang Wang}, \bibinfo{person}{Zhiwen Lin}, {and} \bibinfo{person}{Ke Yan}.} \bibinfo{year}{2024}\natexlab{}.
\newblock \showarticletitle{Vmt-adapter: Parameter-efficient transfer learning for multi-task dense scene understanding}. In \bibinfo{booktitle}{\emph{Proceedings of the AAAI Conference on Artificial Intelligence}}, Vol.~\bibinfo{volume}{38}. \bibinfo{pages}{16085--16093}.
\newblock


\bibitem[Xing et~al\mbox{.}(2023)]%
        {xing2023dual}
\bibfield{author}{\bibinfo{person}{Yinghui Xing}, \bibinfo{person}{Qirui Wu}, \bibinfo{person}{De Cheng}, \bibinfo{person}{Shizhou Zhang}, \bibinfo{person}{Guoqiang Liang}, \bibinfo{person}{Peng Wang}, {and} \bibinfo{person}{Yanning Zhang}.} \bibinfo{year}{2023}\natexlab{}.
\newblock \showarticletitle{Dual modality prompt tuning for vision-language pre-trained model}.
\newblock \bibinfo{journal}{\emph{IEEE Transactions on Multimedia}} (\bibinfo{year}{2023}).
\newblock


\bibitem[Yang et~al\mbox{.}(2023)]%
        {yang2023aim}
\bibfield{author}{\bibinfo{person}{Taojiannan Yang}, \bibinfo{person}{Yi Zhu}, \bibinfo{person}{Yusheng Xie}, \bibinfo{person}{Aston Zhang}, \bibinfo{person}{Chen Chen}, {and} \bibinfo{person}{Mu Li}.} \bibinfo{year}{2023}\natexlab{}.
\newblock \showarticletitle{Aim: Adapting image models for efficient video action recognition}.
\newblock \bibinfo{journal}{\emph{arXiv preprint arXiv:2302.03024}} (\bibinfo{year}{2023}).
\newblock


\bibitem[Yang et~al\mbox{.}(2024)]%
        {yang2024dgl}
\bibfield{author}{\bibinfo{person}{Xiangpeng Yang}, \bibinfo{person}{Linchao Zhu}, \bibinfo{person}{Xiaohan Wang}, {and} \bibinfo{person}{Yi Yang}.} \bibinfo{year}{2024}\natexlab{}.
\newblock \showarticletitle{DGL: Dynamic Global-Local Prompt Tuning for Text-Video Retrieval}. In \bibinfo{booktitle}{\emph{Proceedings of the AAAI Conference on Artificial Intelligence}}, Vol.~\bibinfo{volume}{38}. \bibinfo{pages}{6540--6548}.
\newblock


\bibitem[Yu et~al\mbox{.}(2024)]%
        {yu2024tf}
\bibfield{author}{\bibinfo{person}{Chenyang Yu}, \bibinfo{person}{Xuehu Liu}, \bibinfo{person}{Yingquan Wang}, \bibinfo{person}{Pingping Zhang}, {and} \bibinfo{person}{Huchuan Lu}.} \bibinfo{year}{2024}\natexlab{}.
\newblock \showarticletitle{TF-CLIP: Learning text-free CLIP for video-based person re-identification}. In \bibinfo{booktitle}{\emph{Proceedings of the AAAI Conference on Artificial Intelligence}}, Vol.~\bibinfo{volume}{38}. \bibinfo{pages}{6764--6772}.
\newblock


\bibitem[Zhang(2017)]%
        {zhang2017mixup}
\bibfield{author}{\bibinfo{person}{Hongyi Zhang}.} \bibinfo{year}{2017}\natexlab{}.
\newblock \showarticletitle{mixup: Beyond empirical risk minimization}.
\newblock \bibinfo{journal}{\emph{arXiv preprint arXiv:1710.09412}} (\bibinfo{year}{2017}).
\newblock


\bibitem[Zhang et~al\mbox{.}(2024)]%
        {10744485}
\bibfield{author}{\bibinfo{person}{Junliang Zhang}, \bibinfo{person}{Xu Liu}, \bibinfo{person}{Yu Liang}, \bibinfo{person}{Xiaole Xian}, \bibinfo{person}{Weicheng Xie}, \bibinfo{person}{Linlin Shen}, {and} \bibinfo{person}{Siyang Song}.} \bibinfo{year}{2024}\natexlab{}.
\newblock \showarticletitle{CLIP-Guided Bidirectional Prompt and Semantic Supervision for Dynamic Facial Expression Recognition}. In \bibinfo{booktitle}{\emph{2024 IEEE International Joint Conference on Biometrics (IJCB)}}. \bibinfo{pages}{1--10}.
\newblock
\href{https://doi.org/10.1109/IJCB62174.2024.10744485}{doi:\nolinkurl{10.1109/IJCB62174.2024.10744485}}


\bibitem[Zhang et~al\mbox{.}(2023)]%
        {zhang2023transformer}
\bibfield{author}{\bibinfo{person}{Xiaoqin Zhang}, \bibinfo{person}{Min Li}, \bibinfo{person}{Sheng Lin}, \bibinfo{person}{Hang Xu}, {and} \bibinfo{person}{Guobao Xiao}.} \bibinfo{year}{2023}\natexlab{}.
\newblock \showarticletitle{Transformer-based multimodal emotional perception for dynamic facial expression recognition in the wild}.
\newblock \bibinfo{journal}{\emph{IEEE Transactions on Circuits and Systems for Video Technology}} (\bibinfo{year}{2023}).
\newblock


\bibitem[Zhang et~al\mbox{.}(2014)]%
        {zhang2014bp4d}
\bibfield{author}{\bibinfo{person}{Xing Zhang}, \bibinfo{person}{Lijun Yin}, \bibinfo{person}{Jeffrey~F Cohn}, \bibinfo{person}{Shaun Canavan}, \bibinfo{person}{Michael Reale}, \bibinfo{person}{Andy Horowitz}, \bibinfo{person}{Peng Liu}, {and} \bibinfo{person}{Jeffrey~M Girard}.} \bibinfo{year}{2014}\natexlab{}.
\newblock \showarticletitle{Bp4d-spontaneous: a high-resolution spontaneous 3d dynamic facial expression database}.
\newblock \bibinfo{journal}{\emph{Image and Vision Computing}} \bibinfo{volume}{32}, \bibinfo{number}{10} (\bibinfo{year}{2014}), \bibinfo{pages}{692--706}.
\newblock


\bibitem[Zhao and Liu(2021)]%
        {zhao2021former}
\bibfield{author}{\bibinfo{person}{Zengqun Zhao} {and} \bibinfo{person}{Qingshan Liu}.} \bibinfo{year}{2021}\natexlab{}.
\newblock \showarticletitle{Former-dfer: Dynamic facial expression recognition transformer}. In \bibinfo{booktitle}{\emph{Proceedings of the 29th ACM International Conference on Multimedia}}. \bibinfo{pages}{1553--1561}.
\newblock


\bibitem[Zhao and Patras(2023a)]%
        {zhao2023prompting}
\bibfield{author}{\bibinfo{person}{Zengqun Zhao} {and} \bibinfo{person}{Ioannis Patras}.} \bibinfo{year}{2023}\natexlab{a}.
\newblock \showarticletitle{Prompting visual-language models for dynamic facial expression recognition}.
\newblock \bibinfo{journal}{\emph{arXiv preprint arXiv:2308.13382}} (\bibinfo{year}{2023}).
\newblock


\bibitem[Zhao and Patras(2023b)]%
        {Zhao_2023_BMVC}
\bibfield{author}{\bibinfo{person}{Zengqun Zhao} {and} \bibinfo{person}{Ioannis Patras}.} \bibinfo{year}{2023}\natexlab{b}.
\newblock \showarticletitle{Prompting Visual-Language Models for Dynamic Facial Expression Recognition}. In \bibinfo{booktitle}{\emph{34th British Machine Vision Conference 2023, {BMVC} 2023, Aberdeen, UK, November 20-24, 2023}}. \bibinfo{publisher}{BMVA}.
\newblock
\urldef\tempurl%
\url{https://papers.bmvc2023.org/0098.pdf}
\showURL{%
\tempurl}


\bibitem[Zhou et~al\mbox{.}(2022a)]%
        {zhou2022conditional}
\bibfield{author}{\bibinfo{person}{Kaiyang Zhou}, \bibinfo{person}{Jingkang Yang}, \bibinfo{person}{Chen~Change Loy}, {and} \bibinfo{person}{Ziwei Liu}.} \bibinfo{year}{2022}\natexlab{a}.
\newblock \showarticletitle{Conditional prompt learning for vision-language models}. In \bibinfo{booktitle}{\emph{Proceedings of the IEEE/CVF conference on computer vision and pattern recognition}}. \bibinfo{pages}{16816--16825}.
\newblock


\bibitem[Zhou et~al\mbox{.}(2022b)]%
        {zhou2022learning}
\bibfield{author}{\bibinfo{person}{Kaiyang Zhou}, \bibinfo{person}{Jingkang Yang}, \bibinfo{person}{Chen~Change Loy}, {and} \bibinfo{person}{Ziwei Liu}.} \bibinfo{year}{2022}\natexlab{b}.
\newblock \showarticletitle{Learning to prompt for vision-language models}.
\newblock \bibinfo{journal}{\emph{International Journal of Computer Vision}} \bibinfo{volume}{130}, \bibinfo{number}{9} (\bibinfo{year}{2022}), \bibinfo{pages}{2337--2348}.
\newblock


\bibitem[Zhou et~al\mbox{.}(2024)]%
        {zhou2024dynamic}
\bibfield{author}{\bibinfo{person}{Xin Zhou}, \bibinfo{person}{Dingkang Liang}, \bibinfo{person}{Wei Xu}, \bibinfo{person}{Xingkui Zhu}, \bibinfo{person}{Yihan Xu}, \bibinfo{person}{Zhikang Zou}, {and} \bibinfo{person}{Xiang Bai}.} \bibinfo{year}{2024}\natexlab{}.
\newblock \showarticletitle{Dynamic Adapter Meets Prompt Tuning: Parameter-Efficient Transfer Learning for Point Cloud Analysis}. In \bibinfo{booktitle}{\emph{Proceedings of the IEEE/CVF Conference on Computer Vision and Pattern Recognition}}. \bibinfo{pages}{14707--14717}.
\newblock


\end{thebibliography}










\end{document}